\documentclass[sigconf,nonacm]{acmart}

\usepackage{booktabs}
\usepackage{multirow}
\usepackage{makecell}
\usepackage{graphicx}
\usepackage{xcolor}
\usepackage{pifont}
\usepackage{colortbl}
\definecolor{drop}{HTML}{EC7B6E}
\usepackage{adjustbox}
\usepackage{subcaption}
\usepackage{tikz}
\makeatletter\newcommand{\smallerby}[2]{{\fontsize{\dimexpr#1\dimexpr\f@size pt\relax\relax}{\dimexpr#1\dimexpr\f@size pt\relax\relax}\selectfont #2}}\makeatother

\DeclareRobustCommand{\kM}{\raisebox{-0.15\height}{\includegraphics[height=0.95em]{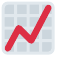}}}\DeclareRobustCommand{\kT}{\raisebox{-0.15\height}{\includegraphics[height=0.95em]{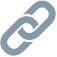}}}\DeclareRobustCommand{\kL}{\raisebox{-0.15\height}{\includegraphics[height=0.95em]{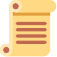}}}\DeclareRobustCommand{\kE}{\raisebox{-0.15\height}{\includegraphics[height=0.95em]{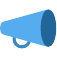}}}

\newcommand{\bench}{RCAtlas}\newcommand{\benchCore}{\bench-Core}

\setcopyright{none}
\renewcommand\footnotetextcopyrightpermission[1]{}

\begin{document}

\setcounter{dbltopnumber}{2}\renewcommand{\dbltopfraction}{0.9}\setcounter{topnumber}{3}\renewcommand{\topfraction}{0.9}\renewcommand{\textfraction}{0.08}\renewcommand{\floatpagefraction}{0.75}\renewcommand{\dblfloatpagefraction}{0.75}
\title[Free to Look, Free to Reason]{Free to Look, Free to Reason: Simple LLM Harnesses Outdiagnose Hand-Crafted Root Cause Analysis Agents}

\author{Tingzhu Bi}
\affiliation{\institution{Peking University}\city{Beijing}\country{China}}
\email{bitingzhu@stu.pku.edu.cn}

\author{Xun Zhang}
\affiliation{\institution{Peking University}\city{Beijing}\country{China}}
\email{zhangxun@stu.pku.edu.cn}

\author{Jinglin Li}
\affiliation{\institution{Beijing University of Posts and Telecommunications}\city{Beijing}\country{China}}
\email{jlli@bupt.edu.cn}

\author{Ping Wang}
\affiliation{\institution{Peking University}\city{Beijing}\country{China}}
\email{pwang@pku.edu.cn}

\author{Meng Ma}
\affiliation{\institution{Peking University}\city{Beijing}\country{China}}
\email{mameng@pku.edu.cn}

\renewcommand{\shortauthors}{Bi et al.}

\begin{abstract}
Root cause analysis (RCA) of microservice failures has attracted many LLM agents whose investigation is hand-crafted, through standard operating procedures, fixed stages and role agents. Each is evaluated on its own data against few baselines, so it is unknown whether this design beats letting a capable model investigate on its own. We compare three simple harnesses that prescribe none of the model's calls (one call over an evidence package, a SQL agent and the Codex coding agent) with five published LLM RCA agents and with statistical and causal methods on \bench{}, 90 validated cases from eight public datasets, across six models and three reasoning efforts. The simple harnesses are the most accurate (PR@1 65--77\% on three mid-tier models), ahead of the published agents (16--53\%) and the non-LLM methods (6--39\%), on every model on which both families were run. The published agents fix from a tenth to all of their model calls in the program, and a case study shows how these prescriptions skip the decisive evidence. A stronger model mainly improves reliability, a multi-turn harness mainly improves aptitude, and more reasoning effort changes accuracy by a few points at most while multiplying the time by up to seven. Among runs more accurate than the best non-LLM method, every Pareto-optimal run in cost or time uses a simple harness.
\end{abstract}
\maketitle

\begin{figure*}[t]
\centering\includegraphics[width=0.74\textwidth]{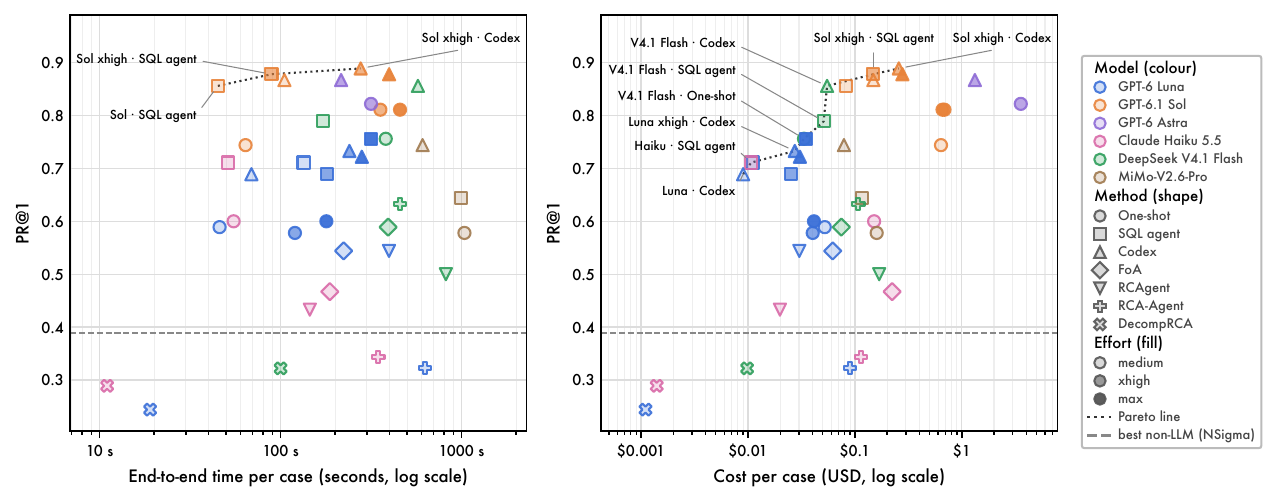}
\caption{Accuracy versus time and cost per case for every run (first pass of each run). Colour = model, shape = method, fill depth = reasoning effort (Luna and Sol). Dashed: PR@1 of the best non-LLM method (NSigma, 38.9\%), a usability threshold fixed in advance; dotted: Pareto line over all runs above it. ChatGPT-login runs priced at OpenRouter list price; time excludes network waits.}
\label{fig:pareto-v2}\end{figure*}

\section{Introduction}\label{sec:intro}

Failures in the microservices behind web services cascade across components, and operators must find the root cause among hundreds of services, pods and nodes from metrics, traces, logs and events~\cite{beyer2016site,zhou2018fault,dang2019aiops}. Three lines of work improve this task: RCA methods, from statistical and causal ranking~\cite{lin2018microscope,wang2018cloudranger,ma2020automap,li2022circa,pham2024baro} to LLM agents; the LLMs underneath them; and general-purpose agent harnesses. A \emph{harness} is the program around a model that feeds it data and carries out its actions. The simplest harness only feeds data, in a single call; richer ones let the model query the data, or write and run code as Codex~\cite{openai_codex} does.

Published LLM RCA agents hand-craft the investigation with standard operating procedures, fixed stages, expert or role sub-agents and votes, so the program decides much of what is read and when~\cite{pei2025foa,wang2024rcagent,xu2025openrca,zhang2024mabc,muhammad2026decomprca}. Each is evaluated against a few baselines on one to three systems, often collected by its own authors, and benchmarks run one reference agent or none~\cite{fang2026openrca2,cai2026agenticopseval}. Whether hand-crafting still pays off as models and harnesses improve is therefore unknown.

We compare the three lines on the same cases. \bench{} pools eight public RCA datasets into one format, screens 542 cases for validity and samples 90 across 23 scenarios. On it we run three simple harnesses that prescribe none of the model's calls (one call over an evidence package, a SQL agent over the raw telemetry, and Codex), five published LLM agents, and statistical and causal methods, with six models, three reasoning efforts and one scorer. We change a method only where it cannot run on our data or has an implementation error.

Harnesses that leave the model free to choose what to look at and how to reason win (Figure~\ref{fig:pareto-v2}). The simple harnesses reach PR@1 65--77\% on three mid-tier models, ahead of the published agents (16--53\%), which are mostly ahead of the non-LLM methods (6--39\%). The trajectories show why: when the program decides what to read, it can skip the decisive evidence, such as a log line or a service that stops receiving calls. A stronger model mainly improves reliability, a multi-turn harness mainly improves aptitude, and more reasoning effort changes little. Among runs more accurate than the best non-LLM method, every Pareto-optimal run in cost or time uses a simple harness.

Our contributions are:
\begin{itemize}
\item \bench{}, 90 validated cases from eight public datasets in one format, with a scoring rule for labels at different levels (Section~\ref{sec:benchmark});
\item a field-wide evaluation of simple harnesses, published LLM agents and non-LLM methods on the same cases, across six models and three reasoning efforts (Section~\ref{sec:exp-overall});
\item analyses of why methods succeed or fail: a case study, the share of model calls each method prescribes, and the separate effects of model capability, reasoning effort and multi-turn investigation on accuracy and reliability (Sections~\ref{sec:exp-case}--\ref{sec:exp-stability});
\item cost- and time-aware recommendations from the Pareto frontier (Section~\ref{sec:exp-pareto}).
\end{itemize}

\section{Related Work}\label{sec:related}

\paragraph{Root cause analysis methods.}
Classical RCA for microservices ranks components from telemetry by statistical deviation (NSigma, BARO~\cite{pham2024baro}, $\epsilon$-Diagnosis~\cite{shan2019ediagnosis}), causal discovery and inference (CIRCA~\cite{li2022circa}, RCD~\cite{ikram2022rcd}, DyCause~\cite{pan2021dycause}, FaultInsight~\cite{bi2024faultinsight}), propagation on dependency graphs (MicroRCA~\cite{wu2020microrca}, MicroRank~\cite{yu2021microrank}), or multi-source fusion (TORAI~\cite{pham2026torai}).
LLM-based methods mostly fix the investigation in the program: standard operating procedures the agent must follow (FoA~\cite{pei2025foa}), a controller that never sees data and an executor that writes code (RCA-Agent~\cite{xu2025openrca}), observation management with expert sub-agents (RCAgent~\cite{wang2024rcagent}), voting among role agents (mABC~\cite{zhang2024mabc}), one agent per trace span (RCLAgent-V2~\cite{rclagentv2}), tree search with self-reflection (LATS-RCA~\cite{naakka2026latsrca}), or program-side retrieval followed by one LLM reranking call (DecompRCA~\cite{muhammad2026decomprca}).
Failure studies trace recurrent errors to the shared agent architecture rather than the model~\cite{kim2026whyfail}, catalogue reasoning failures across agentic and non-agentic prompting~\cite{riddell2026stalled}, and contrast successful and failed trajectories~\cite{lu2026beyond}.

\paragraph{General-purpose agent harnesses.}
Coding agents have grown into general harnesses such as Codex~\cite{openai_codex} and Claude Code~\cite{anthropic_claudecode}: the model decides what to do next and acts by writing and running code, with no task-specific procedure.
For RCA, such a harness without any RCA design already outperforms purpose-built agents on OpenRCA and RCAEval~\cite{huang2026opsharness}.
OpsHarness~\cite{huang2026opsharness} then adds a self-evolving layer that learns from labelled past incidents of the same system.
We take the complementary view: we test simple harnesses without such a layer (Codex, a SQL agent and a single call) against prescribed pipelines and non-LLM methods on the same cases.

\paragraph{Benchmarks and evaluation.}
Model-wide evaluations rank models on many tasks and sub-capabilities; the Artificial Analysis Intelligence Index~\cite{aa_index} has recently added an RCA-style task suite, ITBench-AA~\cite{itbench_aa}, built from ITBench~\cite{jha2025itbench}.
RCA-specific datasets include RCAEval~\cite{pham2025rcaeval}, OpenRCA~\cite{xu2025openrca,fang2026openrca2}, LEMMA-RCA~\cite{zheng2024lemmarca}, the AIOps challenge data and RCA-100~\cite{cai2026agenticopseval,rca100_2026}, and a fault-propagation-aware TrainTicket benchmark~\cite{fang2025rethinking}.
Each covers one to three systems, and each evaluates a narrow slice of methods: OpenRCA~2.0 runs one reference agent across models, AgenticOpsEval~\cite{cai2026agenticopseval} releases data and a scoring protocol without method results, and new methods are often evaluated mainly on data collected by their own authors~\cite{pham2024baro,pham2026torai}.
Audits show that simple rules or telemetry-free priors reach high scores on some of them~\cite{fang2025rethinking,buljic2026graph}.
\bench{} pools eight of these sources, screens every case for validity, and evaluates simple harnesses, published LLM agents and non-LLM methods on the same cases with one scorer.

\paragraph{Agents on complex tasks.}
Evidence from other domains questions elaborate agent pipelines. A fixed localise--repair--validate pipeline competed with software-engineering agents at a fraction of the cost~\cite{xia2024agentless}; splitting a task across conversational turns degrades LLMs mainly through lower reliability~\cite{laban2025lost}; and input length alone can hurt even when retrieval is perfect~\cite{du2025contextlength}. More broadly, general methods that scale with computation tend to overtake hand-engineered knowledge~\cite{sutton2019bitter}. We test this question directly on RCA.

\section{Background and Problem}\label{sec:background}

\paragraph{Systems and telemetry.}
The systems under test are microservice applications on Kubernetes. Their entities are services, the pods and containers that run them, the nodes (hosts) that run the pods, databases and middleware, and the operations a service exposes. Four kinds of telemetry are recorded:
\kM{} \emph{metrics}, time series per entity such as CPU and memory use, request counts, error rates and latency (timestamp, entity, metric name, value);
\kT{} \emph{traces}, one span per call between operations (trace and span identifiers, parent span, operation, duration, status), which reveal who calls whom and where errors and delays arise;
\kL{} \emph{logs}, timestamped lines per pod with a level and a free-text message;
and \kE{} \emph{events}, Kubernetes records such as restarts, evictions or scheduling failures.
Each case also carries an inventory of entities and their relations (calls, deployment, placement).

\paragraph{Task.}
A case gives an incident's flagged period, a period known to precede it, the candidate inventory and the telemetry of both periods. The method returns a ranked list of root-cause entities; we score the first five. Labels name one root-cause entity per case (three cases have two), at the level its dataset annotates: service, pod or node (we count databases as services and hosts as nodes).

\paragraph{Scoring.}
Datasets label at different levels, and an answer may be finer than the label, such as a pod of the faulty service. Before matching, we therefore map each answered entity upward: it is replaced by itself and those of its parents in the inventory (container, pod, service) that lie at the levels the dataset annotates. Duplicates are then dropped, the first five distinct entities form the answer list, and they are compared with the labels by name. For example, when a dataset labels services, an answer naming a pod of cart counts as cart, and a second pod of cart further down the list adds nothing. Methods that rank metric columns, as most non-LLM methods do, are mapped the same way through the entity that owns each column. Each answer can match at most one root cause.

We report two metrics. Let case $i$ of $N$ have the set of root causes $T_i$, with $t_i = |T_i|$, and the mapped answer list $a_{i,1}, \dots, a_{i,k_i}$ ($k_i \le 5$). For a root cause $g \in T_i$, let $r_i(g)$ be the position of the first answer that matches $g$, and $r_i(g) = \infty$ if none does (in particular when the case has no answer). Then
\begin{equation}\label{eq:pr1}
\mathrm{PR@1} = \frac{1}{N} \sum_{i=1}^{N} \mathbf{1}\bigl[\exists\, g \in T_i : r_i(g) = 1\bigr],
\end{equation}
\begin{equation}\label{eq:rr}
\mathrm{RR} = \frac{1}{N} \sum_{i=1}^{N} \frac{1}{t_i} \sum_{g \in T_i} \frac{1}{1 + \max\bigl(0,\, r_i(g) - t_i\bigr)},
\end{equation}
with $1/\infty = 0$. PR@1 is thus the share of cases whose first answer is a root cause. In RR the first $t_i$ positions all score one, so with two root causes A and B both ``A, B'' and ``B, A'' score 1. Both means run over all $N$ cases, so a case without an answer scores zero. We do not report PR@$k$ for larger $k$, because methods return different numbers of answers: the simple harnesses one to five, chosen by the model; RCA-Agent and mABC always one; DecompRCA about 14; and non-LLM methods a full ranking.
To show how hard each case is, we compute its exact random-guess hit rate: the probability that one entity drawn uniformly from the inventory is a hit under the same mapping. Its median over the core set is 3.45\%, and Table~\ref{tab:benchmark} reports its inverse as the effective number of candidates.

\section{The \bench{} Benchmark}\label{sec:benchmark}

\bench{} pools eight public RCA datasets (ten suites, counting RCAEval RE1--3 separately) into one case format, screens every candidate case for validity, and samples the survivors across systems and fault types (Table~\ref{tab:benchmark}; Appendix~\ref{app:cleaning} summarises the construction procedure).

\paragraph{Sources and format.}
The sources span nine systems, from Online Boutique, Sock Shop and TrainTicket to a bank, a telecom system and a cloud platform, with faults injected by chaos tools, by challenge organisers, or in drills on a managed Kubernetes service. Each case is converted to the same tables: an entity inventory with call, deployment and placement relations, and the metrics, traces, logs and Kubernetes events of a window around the fault, all in UTC. The conversion removes traces of the injection tools (for example their pods and log lines) so that no case reveals its answer.

\paragraph{Validity screening.}
The sources have released 2,359 cases; we screened 542 of them, and 348 pass eight exclusion rules. Among others, a case is excluded if its labels contradict each other, if it contains several faults, if the culprit is missing from the data, if neither a judge model nor an automatic check finds evidence of the fault in the telemetry, if data are incomplete, if the period before the fault is not clean, or if the fault starts late in the flagged window. The most frequent reasons are an unclean before-period (69 cases) and no evidence (56).

\paragraph{Sampling.}
From the 348 eligible cases we draw the 90-case core set (\benchCore{}): a minimum per scenario, then cases that diversify fault class, root-cause level, data kinds and telemetry volume. The core set spans 23 scenarios and ten fault classes; the root cause is a service in 59 cases, a pod in 20 and a node in 11. Cases range from 37k to 5.6M telemetry rows, and the exact random-guess hit rate under our scoring rule ranges from 0.25\% to 16.7\% (median 3.45\%). All experiments in this paper use the core set.

\paragraph{Availability.}
We provide the core-set case list and our code at (\url{https://ufile.io/5ww9iyfb}). The file \texttt{cases/core90\_cases.csv} gives, for each case, its identifier in the source dataset, the time window, the data kinds and the answer template, so that every case can be rebuilt from its public source; \texttt{cases/core90\_labels.json} gives the root-cause labels and their levels (59 services, 20 pods and 11 nodes). The folder \texttt{methods/} holds the three simple harnesses (the one-shot method, the SQL agent and the Codex run settings) and our reimplementations of FoA (with its SOP files), RCAgent, DecompRCA and SimpleRCA. Shared code in \texttt{rcatlas/} covers path configuration, an OpenAI-compatible model client, answer parsing and the data tools shared by FoA and RCAgent; \texttt{prompts/} holds the one-shot prompt templates and background notes, and \texttt{answer\_templates/} the answer templates of the 23 scenarios in the version used in the paper, without hints on the root-cause level. We plan to release the converted core-set data (about 11\,GB) and an extended set of about 500 cases built with the same rules.

\begin{table*}[t]
\caption{The \bench{} core set by source. Case pool: all cases the source has released; selected: cases in the core set. \kM{} \kT{} \kL{} \kE: number of selected cases that contain metrics, traces, logs and Kubernetes events. Levels: root causes at service / pod / node level (databases counted as services, hosts as nodes). Rows: metric, trace and log rows in a case's window (median). Candidates: effective number of candidates under our scoring rule, the inverse of the exact random-guess hit rate (median). $^{*}$Fault records of the public release we obtained.}
\label{tab:benchmark}\small\setlength{\tabcolsep}{4pt}
\begin{tabular}{llccccccccc}
\toprule
Source & Systems & Scenarios & Case Pool / Selected & \kM & \kT & \kL & \kE & Levels & Rows & Cand. \\
\midrule
AIOps 2022~\cite{aiops2022} & Online Boutique & 1 & 241$^{*}$ / 6 & 6 & 6 & 6 & -- & 0/3/3 & 650k & 61 \\
AIOps 2025~\cite{cai2026agenticopseval} & Online Boutique, TiDB & 1 & 400 / 9 & 9 & 9 & 9 & -- & 3/3/3 & 251k & 15 \\
ITBench-SRE~\cite{jha2025itbench} & OTel Demo & 1 & 35 / 5 & 5 & 5 & 5 & 5 & 5/0/0 & 457k & 19 \\
LEMMA-RCA~\cite{zheng2024lemmarca} & EKS platform, Book-Info & 2 & 10 / 5 & 5 & -- & 4 & -- & 0/5/0 & 2.1M & 347 \\
OpenRCA 1.0~\cite{xu2025openrca} & Telecom, Bank, Market & 4 & 335 / 14 & 14 & 14 & 10 & -- & 3/9/2 & 571k & 61 \\
OpenRCA 2.0~\cite{fang2026openrca2} & TrainTicket, Hotel, OTel Demo & 4 & 500 / 12 & 12 & 12 & 12 & -- & 12/0/0 & 97k & 25 \\
RCA-100~\cite{rca100_2026,cai2026agenticopseval} & OTel Demo & 1 & 103 / 12 & 12 & 12 & 12 & 12 & 9/0/3 & 1.0M & 34 \\
RCAEval RE1--3~\cite{pham2025rcaeval} & OB, Sock Shop, TrainTicket & 9 & 735 / 27 & 27 & 12 & 18 & -- & 27/0/0 & 681k & 22 \\
\midrule
Total & 8 sources, 10 suites & 23 & 2,359 / 90 & 90 & 70 & 76 & 17 & 59/20/11 & 629k & 29 \\
\bottomrule\end{tabular}\end{table*}

\section{Methods Under Test}\label{sec:methods}

We evaluate three families (Table~\ref{tab:methods}). Every method receives the same information about a case, namely the flagged period, the candidate inventory and the telemetry, and returns a ranked list of root-cause entities, of which the first five are scored.

\paragraph{Simple harnesses.}
\emph{One-shot} sends the whole case to the model in a single call: an evidence package with the system inventory, the metric series that moved (as raw samples), per-minute trace statistics, log templates and events; the model answers with up to five candidates and no tools (Appendix~\ref{app:package}).
The \emph{SQL agent} is a single ReAct loop over the raw files with DuckDB, modelled on the reference agent of OpenRCA~2.0~\cite{fang2026openrca2}: tools to list files, read a schema, run SQL and submit, at most 100 calls.
\emph{Codex} runs the Codex CLI~\cite{openai_codex} in the case folder with a task statement and six rules (read only this folder; no network or installs; do not modify the data; do not try to identify the source dataset; always write a final report; one hour at most); the model writes its own shell and Python.

\paragraph{Published LLM agents.}
We run RCA-Agent~\cite{xu2025openrca} and mABC~\cite{zhang2024mabc} from their released code, and reimplement FoA~\cite{pei2025foa}, RCAgent~\cite{wang2024rcagent} and DecompRCA~\cite{muhammad2026decomprca}, which have no released code, from their papers; our DecompRCA reproduces the paper's numbers on RCAEval RE1. The reimplemented agents share one set of data tools over the raw files.
We follow one rule for fairness: fix only what prevents a method from running on our data or what is an implementation error, and add no capability. Examples are converting time zones for RCA-Agent, restoring modules missing from mABC's release, and completing tool descriptions in FoA's code agent; the appendix summarises the changes.
We exclude RCLAgent-V2~\cite{rclagentv2} (about USD~0.9 and up to an hour per case), LATS-RCA~\cite{naakka2026latsrca} (its released search strips the agents' tools), EviRCA~\cite{wang2026evirca} and MicroRCA-Agent~\cite{tang2025microrcaagent} (each needs a hand-written adapter per system or is built around one benchmark's data layout), and supervised methods, which need a labelled training split.

\paragraph{Non-LLM methods.}
NSigma, BARO, CIRCA, RCD and TORAI come from RCAEval~\cite{pham2025rcaeval}; FaultInsight from the authors' code; SimpleRCA~\cite{fang2025rethinking} is reimplemented from its description. All receive the start of the flagged period; methods with randomness are averaged over three runs. A random ranking gives the floor.

\paragraph{Models and settings.}
Setting~1 runs every LLM method on three mid-tier models: GPT-6 Luna, DeepSeek V4.1 Flash (1017) and Claude Haiku 5.5. Setting~2 runs the simple harnesses on six models, adding GPT-6.1 Sol, GPT-6 Astra (at xhigh) and MiMo-V2.6-Pro. Below we call the models Luna, Flash, Haiku, Sol, Astra and MiMo. Setting~3 raises the reasoning effort of Luna and Sol to xhigh and max. The simple harnesses on Luna, and the SQL agent and Codex on Sol, are run three times. Tables report the mean over passes; Figures~\ref{fig:pareto-v2} and~\ref{fig:behavior} and Table~\ref{tab:costmetrics} use the first pass. No generation parameter is sent except the effort in Setting~3; each case has one hour, and a case without an answer scores zero.

\begin{table}[t]
\caption{Methods under test. Source: original code, original code with interface fixes, or our reimplementation from the paper; the SQL agent follows the reference agent of~\cite{fang2026openrca2}; Model calls decided by: who issues the method's model calls (Section~\ref{sec:exp-prescribed}).}
\label{tab:methods}\small\setlength{\tabcolsep}{5pt}
\begin{tabular}{lll}
\toprule
Method & Source & Model calls decided by \\
\midrule
\multicolumn{3}{l}{\textit{Simple harnesses}} \\
One-shot & ours & one call \\
SQL agent~\cite{fang2026openrca2} & ours & the model \\
Codex~\cite{openai_codex} & Codex CLI & the model \\
\midrule
\multicolumn{3}{l}{\textit{Published LLM agents}} \\
FoA~\cite{pei2025foa} & reimplemented & SOP flow + model \\
RCAgent~\cite{wang2024rcagent} & reimplemented & model + experts \\
RCA-Agent~\cite{xu2025openrca} & original, fixed & controller + executor \\
DecompRCA~\cite{muhammad2026decomprca} & reimplemented & program, one call \\
mABC~\cite{zhang2024mabc} & original, fixed & role agents + voting flow \\
\midrule
\multicolumn{3}{l}{\textit{Non-LLM methods}} \\
NSigma~\cite{pham2024howfar} & RCAEval~\cite{pham2025rcaeval} & -- \\
BARO~\cite{pham2024baro} & RCAEval & -- \\
CIRCA~\cite{li2022circa} & RCAEval & -- \\
RCD~\cite{ikram2022rcd} & RCAEval & -- \\
TORAI~\cite{pham2026torai} & RCAEval & -- \\
FaultInsight~\cite{bi2024faultinsight} & original & -- \\
SimpleRCA~\cite{fang2025rethinking} & reimplemented & -- \\
\bottomrule
\end{tabular}
\end{table}

\section{Experiments}\label{sec:experiments}
We answer five research questions.
\textbf{RQ1} (effectiveness): how do simple harnesses, published LLM agents and non-LLM methods compare on the same cases?
\textbf{RQ2} (mechanism): why do the simple harnesses win, and how much of each method's investigation is prescribed by its program?
\textbf{RQ3} (capability and effort): how do a stronger model and more reasoning effort change how agents investigate and how accurate they are?
\textbf{RQ4} (aptitude and stability): how much accuracy is reproducible across passes, and how much appears only in some?
\textbf{RQ5} (cost-effectiveness): which model and method combinations are Pareto-optimal in accuracy against cost and time?

\subsection{Overall effectiveness (RQ1)}\label{sec:exp-overall}
Table~\ref{tab:main} shows three layers. Averaged over the three mid-tier models, the simple harnesses reach PR@1 65--77\%, the published LLM agents 16--53\%, and the non-LLM methods 6--39\%, against 5.5\% for a random ranking. Codex is best overall, and even the weakest simple harness (one-shot) leads the best published agent (FoA) by 12 points. The lower two layers overlap: NSigma (38.9\%) beats DecompRCA and mABC.
Every LLM method is less accurate on pod- and node-level root causes than on service-level ones, and the published agents lose the most: from service to pod, Codex drops 9 points, FoA and RCAgent 31--38. NSigma is the only method more accurate on pods than on services (55.0\% vs.\ 33.9\%), since resource faults show on the culprit's own metrics. Appendix~\ref{app:details} analyses the hardest groups.
More kinds of data do not make a case easier. The 17 cases with all four kinds, which come from RCA-100 and ITBench-SRE, are the hardest for the simple harnesses (52--71\%), and they form the only column in which a published agent matches the best simple harness: FoA reaches 70.6\%, as Codex does. Conversely, the ten cases with metrics only are the easiest for the SQL agent (95.6\%).
The ordering does not depend on the model (Tables~\ref{tab:models-simple} and~\ref{tab:models-agents}). On each of Luna, Haiku and Flash, every simple harness is ahead of every published agent (one-shot's 5.9-point lead on Luna is not significant), and Codex is best or tied on every model it ran on. Across models, the simple harnesses fall into two tiers at the jump of the AA index from 34.5 to 45.1: 60--71\% on Luna and Haiku, 74--87\% on Flash, Sol and Astra. Leaving out MiMo (see below), 12 of the 13 cross-tier model pairs differ significantly, but only one of the nine within-tier pairs (Appendix~\ref{app:results}). MiMo is the exception: its AA index (46.3) places it in the upper tier, but its accuracy (58--74\%) matches the lower one, partly because it finished only 69--86 of the 90 cases.

\begin{table*}[t]
\caption{Main results on the 90 core cases (\%). LLM methods: mean over Luna, Flash and Haiku. Data kinds: \kM{} metrics, \kT{} traces, \kL{} logs, \kE{} events. Red tint: drop below the column best (white = best, \colorbox{drop!50}{50 points below}, \colorbox{drop!100}{100 points below}). Rel.\ to best: PR@1 as a share of the column best. Random: expected score of a random ordering of the inventory.}
\label{tab:main}
\small\setlength{\tabcolsep}{4.5pt}
\begin{tabular}{lccccccccccc}
\toprule
 & \multicolumn{2}{c}{All (90)} & \multicolumn{5}{c}{PR@1 by data kinds} & \multicolumn{3}{c}{PR@1 by level} & PR@1 \\
\cmidrule(lr){2-3}\cmidrule(lr){4-8}\cmidrule(lr){9-11}
Method & PR@1 & RR & \kM (10) & \kM\kT (4) & \kM\kL (10) & \kM\kT\kL (49) & \kM\kT\kL\kE (17) & service (59) & pod (20) & node (11) & rel.\ to best \\
\specialrule{\lightrulewidth}{\aboverulesep}{0pt}
Codex$^{\dagger}$ & \cellcolor{drop!0}77.2 & \cellcolor{drop!0}80.6 & \cellcolor{drop!11}85.0 & \cellcolor{drop!0}91.7 & \cellcolor{drop!7}86.7 & \cellcolor{drop!0}74.8 & \cellcolor{drop!0}70.6 & \cellcolor{drop!0}81.4 & \cellcolor{drop!0}72.5 & \cellcolor{drop!0}63.6 & 100.0 \\
SQL agent & \cellcolor{drop!4}72.8 & \cellcolor{drop!6}74.9 & \cellcolor{drop!0}95.6 & \cellcolor{drop!11}80.6 & \cellcolor{drop!0}93.3 & \cellcolor{drop!7}67.8 & \cellcolor{drop!10}60.1 & \cellcolor{drop!2}79.7 & \cellcolor{drop!9}63.9 & \cellcolor{drop!11}52.5 & 94.3 \\
One-shot & \cellcolor{drop!12}65.3 & \cellcolor{drop!11}69.8 & \cellcolor{drop!19}76.7 & \cellcolor{drop!8}83.3 & \cellcolor{drop!23}70.0 & \cellcolor{drop!10}65.1 & \cellcolor{drop!18}52.3 & \cellcolor{drop!9}72.5 & \cellcolor{drop!20}52.8 & \cellcolor{drop!14}49.5 & 84.6 \\
\specialrule{\lightrulewidth}{0pt}{0pt}
FoA & \cellcolor{drop!24}53.3 & \cellcolor{drop!22}58.7 & \cellcolor{drop!46}50.0 & \cellcolor{drop!50}41.7 & \cellcolor{drop!40}53.3 & \cellcolor{drop!26}49.0 & \cellcolor{drop!0}70.6 & \cellcolor{drop!15}66.7 & \cellcolor{drop!44}28.3 & \cellcolor{drop!36}27.3 & 69.1 \\
RCAgent & \cellcolor{drop!28}49.3 & \cellcolor{drop!27}53.7 & \cellcolor{drop!39}56.7 & \cellcolor{drop!50}41.7 & \cellcolor{drop!57}36.7 & \cellcolor{drop!24}51.0 & \cellcolor{drop!22}49.0 & \cellcolor{drop!22}59.3 & \cellcolor{drop!44}28.3 & \cellcolor{drop!30}33.3 & 63.8 \\
RCA-Agent & \cellcolor{drop!34}43.3 & \cellcolor{drop!39}41.9 & \cellcolor{drop!62}33.3 & \cellcolor{drop!75}16.7 & \cellcolor{drop!60}33.3 & \cellcolor{drop!25}49.7 & \cellcolor{drop!27}43.1 & \cellcolor{drop!33}48.6 & \cellcolor{drop!42}30.0 & \cellcolor{drop!24}39.4 & 56.1 \\
DecompRCA & \cellcolor{drop!49}28.5 & \cellcolor{drop!46}34.4 & \cellcolor{drop!42}53.3 & \cellcolor{drop!92}0.0 & \cellcolor{drop!77}16.7 & \cellcolor{drop!44}30.6 & \cellcolor{drop!49}21.6 & \cellcolor{drop!50}31.6 & \cellcolor{drop!44}28.3 & \cellcolor{drop!52}12.1 & 36.9 \\
mABC$^{\S}$ & \cellcolor{drop!62}15.6 & \cellcolor{drop!65}16.1 & \cellcolor{drop!91}5.0 & \cellcolor{drop!92}0.0 & \cellcolor{drop!88}5.0 & \cellcolor{drop!60}15.3 & \cellcolor{drop!38}32.4 & \cellcolor{drop!60}21.2 & \cellcolor{drop!65}7.5 & \cellcolor{drop!64}0.0 & 20.1 \\
\specialrule{\lightrulewidth}{0pt}{0pt}
NSigma & \cellcolor{drop!38}38.9 & \cellcolor{drop!33}47.5 & \cellcolor{drop!16}80.0 & \cellcolor{drop!67}25.0 & \cellcolor{drop!43}50.0 & \cellcolor{drop!38}36.7 & \cellcolor{drop!53}17.6 & \cellcolor{drop!47}33.9 & \cellcolor{drop!17}55.0 & \cellcolor{drop!27}36.4 & 50.4 \\
TORAI$^{\ddagger}$ & \cellcolor{drop!44}33.3 & \cellcolor{drop!34}46.8 & \cellcolor{drop!16}80.0 & \cellcolor{drop!67}25.0 & \cellcolor{drop!53}40.0 & \cellcolor{drop!42}32.7 & \cellcolor{drop!65}5.9 & \cellcolor{drop!46}35.6 & \cellcolor{drop!32}40.0 & \cellcolor{drop!55}9.1 & 43.2 \\
SimpleRCA & \cellcolor{drop!44}33.3 & \cellcolor{drop!36}45.1 & \cellcolor{drop!76}20.0 & \cellcolor{drop!42}50.0 & \cellcolor{drop!53}40.0 & \cellcolor{drop!40}34.7 & \cellcolor{drop!41}29.4 & \cellcolor{drop!42}39.0 & \cellcolor{drop!57}15.0 & \cellcolor{drop!27}36.4 & 43.2 \\
BARO & \cellcolor{drop!55}22.2 & \cellcolor{drop!49}31.7 & \cellcolor{drop!56}40.0 & \cellcolor{drop!92}0.0 & \cellcolor{drop!83}10.0 & \cellcolor{drop!50}24.5 & \cellcolor{drop!53}17.6 & \cellcolor{drop!56}25.4 & \cellcolor{drop!52}20.0 & \cellcolor{drop!55}9.1 & 28.8 \\
CIRCA & \cellcolor{drop!59}17.8 & \cellcolor{drop!51}29.2 & \cellcolor{drop!56}40.0 & \cellcolor{drop!92}0.0 & \cellcolor{drop!83}10.0 & \cellcolor{drop!59}16.3 & \cellcolor{drop!53}17.6 & \cellcolor{drop!64}16.9 & \cellcolor{drop!62}10.0 & \cellcolor{drop!27}36.4 & 23.0 \\
FaultInsight$^{\ddagger}$ & \cellcolor{drop!58}18.9 & \cellcolor{drop!51}29.7 & \cellcolor{drop!62}33.3 & \cellcolor{drop!67}25.0 & \cellcolor{drop!77}16.7 & \cellcolor{drop!59}16.3 & \cellcolor{drop!53}17.6 & \cellcolor{drop!57}24.3 & \cellcolor{drop!68}5.0 & \cellcolor{drop!48}15.2 & 24.5 \\
RCD$^{\ddagger}$ & \cellcolor{drop!71}5.9 & \cellcolor{drop!70}10.5 & \cellcolor{drop!96}0.0 & \cellcolor{drop!92}0.0 & \cellcolor{drop!93}0.0 & \cellcolor{drop!68}6.8 & \cellcolor{drop!59}11.8 & \cellcolor{drop!76}5.1 & \cellcolor{drop!69}3.3 & \cellcolor{drop!48}15.2 & 7.7 \\
Random & \cellcolor{drop!72}5.5 & \cellcolor{drop!69}12.0 & \cellcolor{drop!88}7.1 & \cellcolor{drop!89}2.8 & \cellcolor{drop!87}6.6 & \cellcolor{drop!69}5.7 & \cellcolor{drop!67}3.9 & \cellcolor{drop!75}6.3 & \cellcolor{drop!69}3.6 & \cellcolor{drop!59}4.8 & 7.1 \\
\specialrule{\heavyrulewidth}{0pt}{\belowrulesep}\end{tabular}
\par\footnotesize $^{\dagger}$ Haiku not run. $^{\S}$ DeepSeek not run. $^{\ddagger}$ mean over three passes.
\end{table*}

\begin{table*}[t]
\centering
\begin{minipage}[b]{0.6\textwidth}\centering
\caption{The simple harnesses across six models (PR@1 / RR, \%), ordered by the AA Intelligence Index~\cite{aa_index}. Luna and Sol (SQL agent, Codex): mean of three passes. MiMo has no effort setting and finished 69--86 of 90 cases.}
\label{tab:models-simple}
\resizebox{\linewidth}{!}{\small\setlength{\tabcolsep}{4pt}\begin{tabular}[b]{lcccccc}
\toprule
Method & \raisebox{-0.1\height}{\includegraphics[height=1.8em]{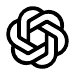}}\,\begin{tabular}[b]{@{}l@{}}\smallerby{0.72}{6}\\[-3pt]\smallerby{0.85}{Luna}\end{tabular} & \raisebox{-0.1\height}{\includegraphics[height=1.8em]{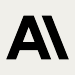}}\,\begin{tabular}[b]{@{}l@{}}\smallerby{0.72}{5.5}\\[-3pt]\smallerby{0.85}{Haiku}\end{tabular} & \raisebox{-0.1\height}{\includegraphics[height=1.8em]{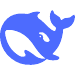}}\,\begin{tabular}[b]{@{}l@{}}\smallerby{0.72}{4.1}\\[-3pt]\smallerby{0.85}{Flash}\end{tabular} & \raisebox{-0.1\height}{\includegraphics[height=1.8em]{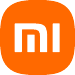}}\,\begin{tabular}[b]{@{}l@{}}\smallerby{0.72}{2.6}\\[-3pt]\smallerby{0.85}{Pro}\end{tabular} & \raisebox{-0.1\height}{\includegraphics[height=1.8em]{logos/openai.pdf}}\,\begin{tabular}[b]{@{}l@{}}\smallerby{0.72}{6.1}\\[-3pt]\smallerby{0.85}{Sol}\end{tabular} & \raisebox{-0.1\height}{\includegraphics[height=1.8em]{logos/openai.pdf}}\,\begin{tabular}[b]{@{}l@{}}\smallerby{0.72}{6}\\[-3pt]\smallerby{0.85}{Astra}\end{tabular} \\
\textit{AA index} & 29.9 & 34.5 & 45.1 & 46.3 & 47.8 & 52.4 \\
\textit{Effort} & medium & medium & default & -- & medium & xhigh \\
\midrule
SQL agent & 68.5 / 69.5 & 71.1 / 74.5 & 78.9 / 80.6 & 64.4 / 66.1 & 84.1 / 82.8 & -- \\
Codex & 68.9 / 73.1 & -- & 85.6 / 88.1 & 74.4 / 77.6 & 87.0 / 88.2 & 86.7 / 88.3 \\
One-shot & 60.4 / 64.4 & 60.0 / 66.0 & 75.6 / 79.1 & 57.8 / 63.6 & 74.4 / 77.7 & 82.2 / 84.3 \\
\bottomrule\end{tabular}}
\end{minipage}\hfill
\begin{minipage}[b]{0.37\textwidth}\centering
\caption{The published LLM agents on the three mid-tier models (PR@1 / RR, \%; header as in Table~\ref{tab:models-simple}). mABC: partial runs (47 and 85 of 90 cases answered).}
\label{tab:models-agents}
\resizebox{\linewidth}{!}{\small\setlength{\tabcolsep}{4pt}\begin{tabular}[b]{lccc}
\toprule
Method & \raisebox{-0.1\height}{\includegraphics[height=1.8em]{logos/openai.pdf}}\,\begin{tabular}[b]{@{}l@{}}\smallerby{0.72}{6}\\[-3pt]\smallerby{0.85}{Luna}\end{tabular} & \raisebox{-0.1\height}{\includegraphics[height=1.8em]{logos/anthropic.pdf}}\,\begin{tabular}[b]{@{}l@{}}\smallerby{0.72}{5.5}\\[-3pt]\smallerby{0.85}{Haiku}\end{tabular} & \raisebox{-0.1\height}{\includegraphics[height=1.8em]{logos/deepseek.pdf}}\,\begin{tabular}[b]{@{}l@{}}\smallerby{0.72}{4.1}\\[-3pt]\smallerby{0.85}{Flash}\end{tabular} \\
\textit{AA index} & 29.9 & 34.5 & 45.1 \\
\textit{Effort} & medium & medium & default \\
\midrule
FoA & 54.4 / 57.1 & 46.7 / 56.7 & 58.9 / 62.2 \\
RCAgent & 54.4 / 57.0 & 43.3 / 48.4 & 50.0 / 55.6 \\
RCA-Agent & 32.2 / 30.6 & 34.4 / 33.3 & 63.3 / 61.7 \\
DecompRCA & 24.4 / 30.4 & 28.9 / 35.9 & 32.2 / 37.0 \\
mABC & 11.1 / 11.7 & 20.0 / 20.6 & -- \\
\bottomrule\end{tabular}}
\end{minipage}
\end{table*}

\subsection{Why simple harnesses win: a case study (RQ2)}\label{sec:exp-case}

\begin{figure}[t]
\centering\includegraphics[width=0.78\columnwidth]{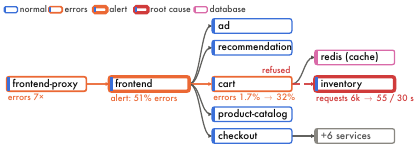}
\caption{Case rca100-main-0104. Orange: calls on the failing path; dashed red: the refused call.}
\label{fig:case-topology}
\end{figure}
\begin{figure}[t]
\centering\includegraphics[width=0.78\columnwidth]{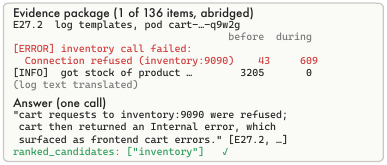}
\caption{One-shot (Luna, pass 1): the evidence item the answer cites, and the answer.}
\label{fig:case-oneshot}
\end{figure}
\begin{figure*}[t]
\centering

\resizebox{\textwidth}{!}{
\adjustbox{valign=t}{\includegraphics{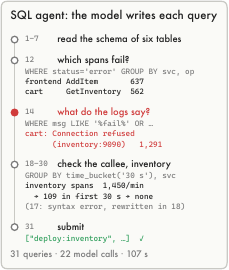}}\hspace{5pt}
\adjustbox{valign=t}{\includegraphics{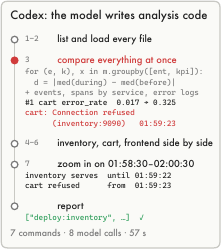}}\hspace{5pt}
\adjustbox{valign=t}{\includegraphics{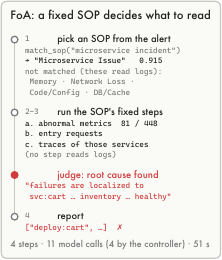}}\hspace{5pt}
\adjustbox{valign=t}{\includegraphics{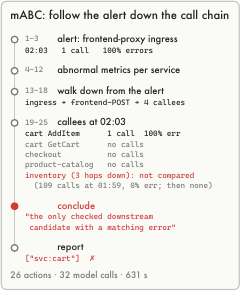}}}
\caption{Four runs on case rca100-main-0104 (Luna). Grey: what the model wrote; black: what came back; red dot: the turning point. Times as recorded, including network waits.}
\label{fig:case-runs}
\end{figure*}
The difference between the families is who decides what to look at and how to compute on it, which is what the prescribed-call ratio (Section~\ref{sec:exp-prescribed}) measures. The SQL agent writes its own queries and Codex writes its own analysis code, so the model chooses both what to read and how to compare it, and none of their calls is prescribed. The published agents fix part of this in the program. FoA can only choose among the steps of its built-in procedures, so when the matched procedure has no log step, it never sees the logs. RCA-Agent's controller never touches the data and learns about the case only from its executor's summaries, and mABC starts from an alert its program computes and walks the call chain from there. In the terms of our title, the SQL agent and Codex are free to look and free to reason; one-shot is free to reason over evidence that rules select.

We follow one case through the methods, rca100-main-0104 from RCA-100 (Figure~\ref{fig:case-topology}). The inventory Service selects no pods, so cart's calls to it are refused. The root cause raises no errors of its own: its request count collapses, while cart's error rate rises and cart logs ``Connection refused (inventory:9090)''. All runs below use Luna.

\paragraph{One-shot.} The program decides what the model reads (the evidence-package rules), and the model does the reasoning. The package holds a log template from the cart pod (E27.2) whose refusal line grows from 43 lines before the fault to 609 during it. All three passes cite this item and rank inventory first (Figure~\ref{fig:case-oneshot}).

\paragraph{SQL agent.} The SQL agent starts from the symptom and follows ``who cannot reach whom'' to the root cause (Figure~\ref{fig:case-runs}, left). Its 31 queries reduce to six steps: it reads which tables exist (steps 1--7); counts failing calls by service and finds cart failing (step 12); searches the logs for error keywords and finds 1,291 lines in which cart's connections to inventory are refused (step 14); queries inventory's requests per minute and confirms that they fall from about 1,450 to 109 in the first 30 seconds of the fault and then to none; checks inventory's pod logs and its place in the topology (steps 18--30, including an SQL error at step 17 that it fixes itself); and submits inventory (step 31).

\paragraph{Codex.} Codex casts a wide net first and then narrows down, in seven commands. After listing and loading the files, its third command runs comparison code it wrote itself over every entity and metric before and during the fault, together with events, spans and error logs; cart's error rate (0.017 to 0.325) ranks first, and the refusal line is already in the output. It then pulls the events and the metrics and logs of inventory, cart and frontend side by side (commands 4--6), reads the logs of the two minutes around the onset (command 7), which show inventory serving until 01:59:22 and cart refused from 01:59:23, and writes its report naming inventory.

\paragraph{FoA.} The SOP chosen by the program decides what is read. FoA fills in the fault description itself as ``microservice incident'', which matches a generic SOP whose fixed steps never read logs; its judge declares the root cause found, cart, in the first round, after four calls of the controller (11 model calls in all). In the original paper the alert carries the fault type (``Network Issue'' in its Figure~4), from which the SOP is chosen; we give every method only the anomalous period, because the fault type would reveal the answer. Four of our 12 SOPs, written after the paper's examples, read logs, but none matched this case (the SOP files are in the released code; Appendix~\ref{app:walkthroughs}).

\paragraph{mABC.} The framework walks down the call chain from the alert: the frontend-proxy ingress at 02:03, frontend's POST, and its four callees. In that minute only cart's AddItem still receives a call (one, failing); the other three receive none. Inventory, three hops down, is never compared at the interface level; it served 109 calls with no errors at 01:59 and none afterwards. mABC stops at cart. A method that walks the call chain and inspects only the calls a service receives cannot see a service that stops receiving them.

\paragraph{Non-LLM methods.} NSigma and BARO score each metric by its largest standardised deviation, keeping the sign, so only metrics that rise score high. Inventory's request count falls sharply; its z-score is large and negative, and the metric ranks second to last among the 373 columns.

\subsection{How much of the investigation is prescribed (RQ2)}\label{sec:exp-prescribed}

To measure what the case study shows across all cases, we compute each method's \emph{prescribed-call ratio}: the share of its model calls that its program issues at fixed points rather than the model choosing to make them, over the trajectories of all 90 cases on the three mid-tier models (Table~\ref{tab:freedom}).
From the run records and each method's code, we sort every model call into one of three kinds:
\begin{itemize}
\item \emph{main}: the next turn of the model in charge of the task, such as every call of Codex and the SQL agent, the main agent of FoA, and the controllers of RCAgent and RCA-Agent;
\item \emph{delegated}: a call the main model chooses to hand to another role, such as FoA's code agent or RCA-Agent's executor writing code;
\item \emph{prescribed}: a call the program makes whatever the model decides, such as FoA's action agent at every step and its judge after every SOP run, RCAgent's chunked analysis and summary calls, RCA-Agent's summary after each successful execution, forced final reports, and DecompRCA's reranking of 15 candidates selected by its program.
\end{itemize}
For a method run on model $m$, let $C_m$ be the cases with a trajectory, and let case $i$ have $c_i$ model calls, of which $f_i$ are prescribed. The ratio is averaged over cases and then over the three models $\mathcal{M}$:
\begin{equation}\label{eq:presc}
\rho = \frac{1}{|\mathcal{M}|} \sum_{m \in \mathcal{M}} \frac{1}{|C_m|} \sum_{i \in C_m} \frac{f_i}{c_i}.
\end{equation}

The three simple harnesses prescribe no call. Every published agent does: at least 0.10 for mABC, 0.29 for RCA-Agent, 0.35 for RCAgent, 0.51 for FoA, and 1.00 for DecompRCA, whose single call is placed by its program. Within the published agents the ratio does not order accuracy (RCA-Agent prescribes fewer calls than FoA yet is less accurate); the clear divide is between harnesses that prescribe nothing and agents that prescribe something.

Three points qualify these numbers. First, the one call of one-shot counts as a main call, because the model receives the whole task, whereas DecompRCA's single call counts as prescribed, because the model only reranks a list the program has already narrowed. One-shot's evidence package is still selected by rules (it keeps only metric series that moved), which is why the case study says the program decides what one-shot reads. Second, mABC's records do not say which role made each call, so its ratio is a lower bound. Third, simply counting calls does not separate the families: the SQL agent makes 27 calls per case, the same order as the multi-step published agents (35--50), while one-shot and DecompRCA both make one.

\begin{table}[!htbp]
\caption{Prescribed-call ratio (Equation~\ref{eq:presc}) on all 90 cases and the three mid-tier models. Calls, actions: per case; per-case range: lowest and highest value over all cases and models; fixed stages: rules or stages read from the code. mABC: lower bound.}
\label{tab:freedom}\small\setlength{\tabcolsep}{3.5pt}
\begin{tabular}{lrrrcr}
\toprule
Method & \shortstack[r]{Calls\\/ case} & \shortstack[r]{Actions\\/ case} & \shortstack[r]{Prescribed-\\call ratio} & \shortstack{Per-case\\range} & \shortstack[r]{Fixed\\stages} \\
\midrule
Codex & 8.5 & 7.5 & 0.00 & 0.00--0.00 & 0 \\
SQL agent & 26.8 & 41.7 & 0.00 & 0.00--0.08 & 1 \\
One-shot & 1.0 & 0.0 & 0.00 & -- & 0 \\
\midrule
FoA & 36.4 & 22.3 & 0.51 & 0.46--0.55 & 8 \\
RCAgent & 35.1 & 21.5 & 0.35 & 0.05--0.81 & 4 \\
RCA-Agent & 50.3 & 32.3 & 0.29 & 0.06--0.33 & 4 \\
mABC & 41.2 & 35.9 & $\geq$0.10 & -- & 3 \\
DecompRCA & 1.0 & 2.0 & 1.00 & 1.00--1.00 & 2 \\
\bottomrule\end{tabular}\end{table}

\subsection{Model capability and reasoning effort (RQ3)}\label{sec:exp-capability}
\begin{figure*}[tbp]
\centering
\setbox0=\hbox{\footnotesize\setlength{\tabcolsep}{2.5pt}\begin{tabular}{clccccc}
\toprule
Contrast & Method & Calls & Actions & \shortstack{Entities\\coverage} & \shortstack{RC\\visited} & $\Delta$PR@1 \\
\midrule
A & Codex & 8.5$\rightarrow$7.5 & 7.5$\rightarrow$6.1 & 16\%$\rightarrow$12\% & 89\%$\rightarrow$94\% & \textbf{+17.8} \\
 & SQL & 19.7$\rightarrow$8.9 & 31.4$\rightarrow$24.9 & 21\%$\rightarrow$12\% & 91\%$\rightarrow$93\% & \textbf{+14.4} \\
\midrule
B & Codex & 8.5$\rightarrow$13.3 & 7.5$\rightarrow$12.2 & 16\%$\rightarrow$25\% & 89\%$\rightarrow$94\% & \textbf{+3.3} \\
 & SQL & 19.7$\rightarrow$24.3 & 31.4$\rightarrow$53.9 & 21\%$\rightarrow$27\% & 91\%$\rightarrow$96\% & \textbf{+4.4} \\
\midrule
C & Codex & 7.5$\rightarrow$9.2 & 6.1$\rightarrow$8.5 & 12\%$\rightarrow$17\% & 94\%$\rightarrow$99\% & \textbf{+2.2} \\
 & SQL & 8.9$\rightarrow$10.3 & 24.9$\rightarrow$33.8 & 12\%$\rightarrow$18\% & 93\%$\rightarrow$96\% & \textbf{+2.2} \\
\bottomrule\end{tabular}}
$\vcenter{\hbox{\includegraphics{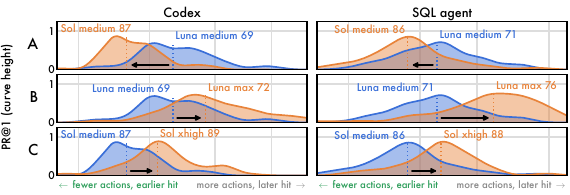}}}$\hfill$\vcenter{\box0}$
\caption{Stronger model versus more reasoning effort (90 cases, first pass). Contrasts: A = Luna $\rightarrow$ Sol (medium); B = Luna medium $\rightarrow$ max; C = Sol medium $\rightarrow$ xhigh. Left: cases placed by the geometric mean of tool actions and the step at which the root cause is first queried (log scale; left = fewer, earlier); curve height = the run's PR@1; dotted line = median; arrow = shift from blue to orange. Right: per-case means; Entities coverage = share of inventory entities queried; RC visited = share of cases whose queries ever name the root cause.}
\label{fig:behavior}\end{figure*}

Because the simple harnesses leave the investigation to the model, their accuracy rests on the model: it rises with the model's tier (Section~\ref{sec:exp-overall}). We now ask how a stronger model and more reasoning effort change the investigation itself. Figure~\ref{fig:behavior} compares three pairs of runs. A stronger model investigates less and finds more (contrast A): from Luna to Sol, both agents make fewer calls and query fewer entities (Codex 16\% $\rightarrow$ 12\%, SQL agent 21\% $\rightarrow$ 12\%), yet reach the root cause in more cases and gain 14--18 points of PR@1.
More reasoning effort does the opposite (contrasts B and C, Luna medium to max and Sol medium to xhigh): agents make more calls and query more entities, but gain only 2--4 points of PR@1 (first pass of the SQL agent and Codex), at 1.5 to 4 times the wall time. Appendix~\ref{app:results} covers every effort level (Figure~\ref{fig:effort}). In all three contrasts the runs query the root cause in two to five more cases out of a hundred (RC visited); the difference is what the model makes of it. Sol turns what it sees into a correct first answer, while a larger budget mostly adds queries that the model then does not use to change its answer. A larger budget buys exploration; a stronger model buys direction.

\subsection{Aptitude and stability (RQ4)}\label{sec:exp-stability}

\begin{figure}[tbp]
\centering\includegraphics[width=0.66\columnwidth]{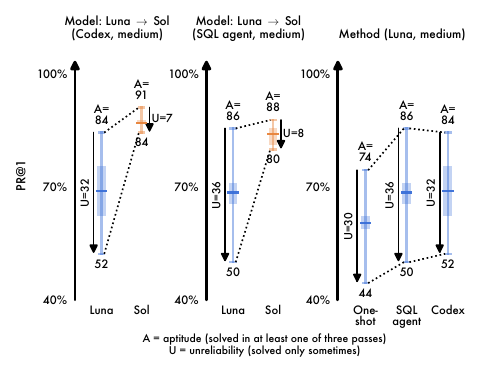}
\caption{Aptitude (A) and unreliability (U) over three passes (PR@1, medium effort). Whiskers: A$-$U to A; box: range of the three passes; line: their mean.}
\label{fig:aptitude}
\end{figure}

The comparisons so far use one pass per run, but an LLM's answers vary between passes. Following~\citet{laban2025lost}, we run Luna and Sol three times at medium effort and measure aptitude and unreliability. \citet{laban2025lost} define them through the 90th and 10th percentiles of a score over many samples; with three passes we use the best and the worst pass. Let $s_i^{p} \in \{0, 1\}$ be the PR@1 of case $i$ in pass $p \in \{1, 2, 3\}$. Then
\begin{equation}\label{eq:apt}
A = \frac{1}{N} \sum_{i=1}^{N} \max_{p} s_i^{p}, \qquad
U = \frac{1}{N} \sum_{i=1}^{N} \Bigl(\max_{p} s_i^{p} - \min_{p} s_i^{p}\Bigr),
\end{equation}
so $A$ is the share of cases solved in at least one pass, $U$ the share solved in some passes but not all, and $A - U$ the share solved in all three. Figure~\ref{fig:aptitude} shows both. A stronger model mainly buys reliability: from Luna to Sol, A rises by only 2--7 points, but U falls from 32--36 to 7--8. On Luna, a single pass of Codex ranges from 62\% to 76\%. A multi-turn agent mainly buys aptitude: on Luna, the SQL agent and Codex solve 10--11 points more cases at least once than one-shot, with similar U (30--36). One-shot never solves 26\% of the cases in any pass, against 14--16\% for the two agents, so it misses cases outright rather than by chance. This is the opposite of the finding of~\citet{laban2025lost}, where splitting a task over turns mainly costs reliability.
Single passes on a weaker model are therefore noisy estimates. On Luna the three passes of Codex give 62\%, 69\% and 76\%, and those of the SQL agent 66\% to 71\%, so a comparison between methods based on one pass of a weak model can flip. On Sol the passes differ by at most 4.5 points (Codex 86.7--87.8\%, SQL agent 81.1--85.6\%).

\subsection{Cost-effectiveness (RQ5)}\label{sec:exp-pareto}
In operations, accuracy is not the only concern: an incident needs an answer quickly and at a sustainable cost. Figure~\ref{fig:pareto-v2} plots PR@1 against time and against cost per case. We count a run as usable only if it is more accurate than the best non-LLM method (NSigma, PR@1 38.9\%), a threshold fixed before drawing the frontier, and draw the Pareto lines through all usable runs. Seven runs of published agents qualify (FoA, RCAgent and RCA-Agent, 43--63\%), yet every run on either line uses a simple harness: each published agent is beaten in accuracy by a simple harness that is no more costly, and by one that is no slower. The DecompRCA runs, the cheapest and fastest of all, fall below the threshold (24--29\%).
The frontier gives three practical choices. When cost matters most, Luna with Codex reaches 68.9\% for under one cent per case. When accuracy per dollar matters, Flash with Codex reaches 85.6\% for about five cents. When the answer is needed quickly, Sol with the SQL agent answers in 45\,s at 85.6\%. The most accurate run, Sol at xhigh with Codex (88.9\%), costs about 26 cents and takes 277\,s per case, so the last few points of accuracy cost roughly five times as much as the cheapest strong setup.

\section{Limitations}\label{sec:limitations}

Our study has five limitations. First, Codex is the only general-purpose coding harness we test; others, such as Claude Code, may behave differently. Second, all models are from 2026, so we cannot say whether the ordering holds for the weaker models for which many prescribed workflows were designed. Third, budget kept us from current frontier models such as Fable, Opus or Argon. Fourth, despite careful conversion and reimplementation, our datasets and methods may still differ from their originals; the appendix summarises the changes. Fifth, the case study follows a single case. The prescribed-call ratio extends its mechanism to all 90 cases, but it measures how much of an investigation is fixed, not that this fixing causes the lower accuracy; an ablation that lifts the prescriptions from one agent would test that directly.

\section{Conclusion}\label{sec:conclusion}
For microservice root cause analysis, a capable model that is given the data and left to decide what to look at and how to reason outdiagnoses agents whose investigation is hand-crafted, and both mostly outperform statistical and causal methods. A stronger model improves accuracy and reliability; more reasoning effort adds little. Practitioners can start from a simple harness and pick the model and harness on the cost--accuracy frontier. Designers of RCA agents should compare against such harnesses on shared, validated cases, which \bench{} provides.

Our results do not show that domain knowledge is useless. They show that fixing it into the control flow hurts when a procedure does not fit the case, as in our case study, where the procedure that FoA matched never read the logs. Offering procedures to the model as optional guidance, rather than prescribing them, is a natural next step.

\appendix
\setcounter{dbltopnumber}{2}\renewcommand{\dbltopfraction}{0.95}\renewcommand{\textfraction}{0.05}

\section{Per-method walkthroughs}\label{app:walkthroughs}
We explain why the methods not covered by the case study (Section~\ref{sec:exp-case}) succeed or fail; PR@1 is given as Luna / Flash / Haiku.

\paragraph{Published LLM agents.}
\emph{FoA} (54.4 / 58.9 / 46.7) runs one of twelve SOPs written before any run after the examples in~\cite{pei2025foa}. In FoA the alert names the fault type, which selects the SOP; our setting withholds it, since it would reveal the answer, so FoA describes the incident itself. Luna then matches the generic \emph{Microservice Issue} SOP (abnormal metrics $\rightarrow$ entry requests $\rightarrow$ traces) in 69 of the 90 cases; Flash, which more often guesses a fault type, does so in 33. Only four SOPs contain a log step (Memory, Network Loss or Partition, Code or Configuration Error, Database or Cache). FoA's accuracy varies widely by fault group: on Luna 93\% on application-code faults but 41\% on resource saturation.
\emph{RCAgent} (54.4 / 50.0 / 43.3) prescribes 35\% of its calls and, with FoA, comes closest to the simple harnesses on Luna; Luna leaves 7 cases and Flash 19 cases unanswered within its 30 steps.
\emph{RCA-Agent} (32.2 / 63.3 / 34.4): its released executor discards outputs above 16,384 tokens after two tries, which costs Luna, which prints whole tables, 16\% of its executor steps (Haiku 3\%, yet Haiku scores as low).
\emph{DecompRCA} (24.4 / 32.2 / 28.9): the program's at most 15 candidates contain the root cause in only 57.8\% of the cases, which bounds the model's rerank.
\emph{mABC} (11.1 / -- / 20.0) walks down the call chain from the alert, so a service that receives no requests is never reached. Luna also declares the cause undetermined in 42 of 89 cases, concentrated in sources with few or no traces (LEMMA 5 of 5, RCAEval RE1 8 of 9); Haiku answers 85 of 90.

\paragraph{Non-LLM methods.}
The gap lies in where the evidence is. On resource saturation (34 cases) NSigma reaches 62\%, close to one-shot on Luna (65\%); on errors, outages, configuration faults and network breaks (33 cases), whose evidence is in logs, traces and events, NSigma and BARO reach 15\% and 6\% against 67\% for the SQL agent and Codex on Luna. Three causes add to this. Many sources have thousands of series with about 15 pre-fault samples, half of them constant, so normal jumps outrank the root cause. NSigma and BARO score the largest \emph{signed} deviation, so a drop ranks last (inventory's request drop in the case study: 372nd of 373). Causal-graph methods cannot learn dependencies from tens of samples: RCD misses the root cause entirely in 71 of 89 cases.

\section{Additional results and significance}\label{app:results}
\paragraph{Significance.} A paired bootstrap over the 90 cases (10,000 resamples, 95\%) confirms the main gaps: the SQL agent and Codex beat the best published agent on every model they ran on, by 14 to 24 points (one-shot on Flash and Haiku only), and, with MiMo left out of the tiers as in Section~\ref{sec:exp-overall}, 12 of 13 cross-tier model pairs differ against 1 of 9 within-tier pairs (with MiMo grouped with the lower tier by its accuracy: 20 of 21 and 1 of 14). The best published agent beats NSigma on Luna and Flash, but mABC falls significantly below it, so the family order holds for the strongest members.

\begin{figure}[t]
\centering\includegraphics[width=0.9\columnwidth]{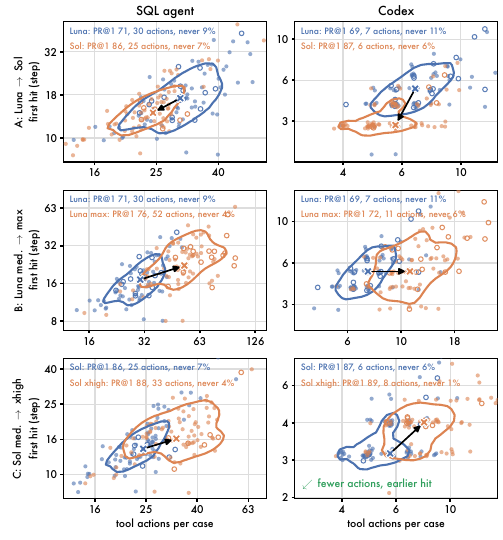}
\caption{The cases behind Figure~\ref{fig:behavior} (first pass). x = tool actions per case, y = step of the first query that names the root cause (both log); filled = root cause ranked first, hollow = missed; contour = densest 65\% of a run's hits; X = median; arrow = first run (blue) to second (orange). Cases that never query the root cause are not drawn.}
\label{fig:orientation}
\end{figure}

\paragraph{Investigation orientation (Figure~\ref{fig:orientation}).} Figure~\ref{fig:behavior} summarises each run along one diagonal; Figure~\ref{fig:orientation} shows the cases behind it. From Luna to Sol (A) both clouds move down and left: the SQL agent needs 25 instead of 30 actions and first queries the root cause at step 14 instead of 17, Codex at step 3 instead of 5. More effort (B, C) moves the clouds right without moving them down: Luna at max uses 52 instead of 30 actions with the SQL agent and first hits the root cause later (step 22), and Sol at xhigh hits it no earlier than at medium.

\paragraph{By source (Table~\ref{tab:bysource}).} The simple harnesses lead on seven of eight sources. On RCA-100 FoA leads (72.2), although Luna's FoA runs used the generic SOP in 10 of its 12 cases. Non-LLM methods come closest on RCAEval, OpenRCA~1.0 and LEMMA, where NSigma reaches 57--60\% and on OpenRCA~1.0 even exceeds one-shot (57.1 vs.\ 56.3); AIOps~2022 is the hardest source for every method (best 52.8).

\paragraph{Cost (Table~\ref{tab:costmetrics}).} The table reports the first pass of each run, whereas Tables~\ref{tab:main}--\ref{tab:models-agents} average three passes for Luna and Sol, so the two can differ slightly. One-shot sends 235k--293k prompt tokens in one call. The agents' totals range from 90k (Sol with the SQL agent) to over two million (Flash and MiMo), but 76--95\% of them are served from the provider's cache, so the SQL agent on Luna costs USD~0.011 per case against 0.052 for one-shot. Weaker models take more steps: Flash and MiMo make 21--47 calls per case with the agents, Luna and Sol 7.5--20.

\begin{figure}[t]
\centering\includegraphics[width=0.60\columnwidth]{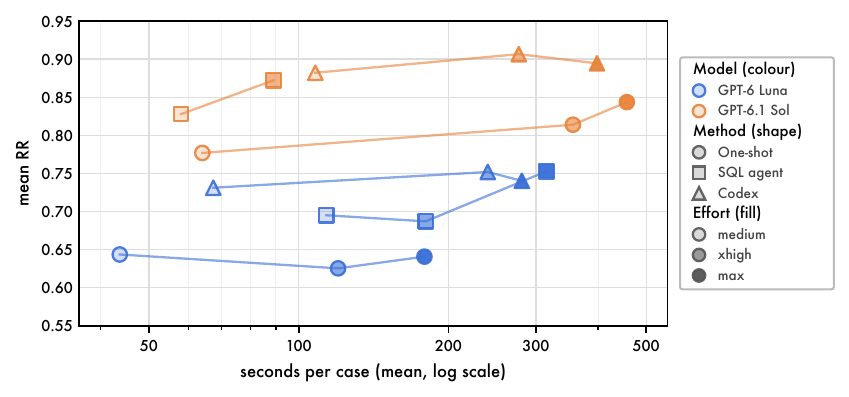}
\caption{RR versus wall time per case (without network waits) as reasoning effort rises from medium to xhigh and max. Colour = model, shape = method, fill depth = effort. Medium points of Luna and of Sol's SQL agent and Codex: mean of three passes; others one pass. Sol's SQL agent was not run at max.}
\label{fig:effort}\end{figure}

\paragraph{Reasoning effort (Figure~\ref{fig:effort}).} Over every effort level and harness in Figure~\ref{fig:effort}, including one-shot at max (which Section~\ref{sec:exp-capability} does not cover), raising effort above medium multiplies wall time by 1.5 to 7 and changes RR by $-$2 to $+$7 points, while replacing Luna with Sol at medium raises it by 13 to 15 points: a stronger model is worth more than a larger thinking budget.

\begin{table*}[t]
\caption{Results by source dataset (\%), averaged as in Table~\ref{tab:main}. Red tint: drop below the column best (white = best). Rel.\ to best: PR@1 as a share of the column best.}
\label{tab:bysource}
\footnotesize\setlength{\tabcolsep}{4pt}
\begin{tabular}{lccccccccccc}
\toprule
 & \multicolumn{2}{c}{All (90)} & \multicolumn{8}{c}{PR@1 by source dataset} & PR@1 \\
\cmidrule(lr){2-3}\cmidrule(lr){4-11}
Method & PR@1 & RR & \shortstack{RCAEval\\(27)} & \shortstack{OpenRCA\\1.0\\(14)} & \shortstack{OpenRCA\\2.0\\(12)} & \shortstack{RCA-100\\(12)} & \shortstack{AIOps\\2025\\(9)} & \shortstack{AIOps\\2022\\(6)} & \shortstack{ITBench\\(5)} & \shortstack{LEMMA\\(5)} & rel.\ to best \\
\specialrule{\lightrulewidth}{\aboverulesep}{0pt}
Codex$^{\dagger}$ & \cellcolor{drop!0}77.2 & \cellcolor{drop!0}80.6 & \cellcolor{drop!4}90.1 & \cellcolor{drop!0}65.5 & \cellcolor{drop!2}72.2 & \cellcolor{drop!6}66.7 & \cellcolor{drop!0}81.5 & \cellcolor{drop!0}52.8 & \cellcolor{drop!4}80.0 & \cellcolor{drop!0}96.7 & 100.0 \\
SQL agent & \cellcolor{drop!4}72.8 & \cellcolor{drop!6}74.9 & \cellcolor{drop!0}94.2 & \cellcolor{drop!4}61.1 & \cellcolor{drop!0}74.1 & \cellcolor{drop!22}50.0 & \cellcolor{drop!16}65.4 & \cellcolor{drop!23}29.6 & \cellcolor{drop!0}84.4 & \cellcolor{drop!1}95.6 & 94.3 \\
One-shot & \cellcolor{drop!12}65.3 & \cellcolor{drop!11}69.8 & \cellcolor{drop!14}79.8 & \cellcolor{drop!9}56.3 & \cellcolor{drop!2}72.2 & \cellcolor{drop!22}50.0 & \cellcolor{drop!15}66.7 & \cellcolor{drop!10}42.6 & \cellcolor{drop!27}57.8 & \cellcolor{drop!32}64.4 & 84.6 \\
\specialrule{\lightrulewidth}{0pt}{0pt}
FoA & \cellcolor{drop!24}53.3 & \cellcolor{drop!22}58.7 & \cellcolor{drop!28}66.7 & \cellcolor{drop!37}28.6 & \cellcolor{drop!7}66.7 & \cellcolor{drop!0}72.2 & \cellcolor{drop!44}37.0 & \cellcolor{drop!47}5.6 & \cellcolor{drop!18}66.7 & \cellcolor{drop!50}46.7 & 69.1 \\
RCAgent & \cellcolor{drop!28}49.3 & \cellcolor{drop!27}53.7 & \cellcolor{drop!35}59.3 & \cellcolor{drop!32}33.3 & \cellcolor{drop!10}63.9 & \cellcolor{drop!19}52.8 & \cellcolor{drop!30}51.9 & \cellcolor{drop!31}22.2 & \cellcolor{drop!44}40.0 & \cellcolor{drop!63}33.3 & 63.8 \\
RCA-Agent & \cellcolor{drop!34}43.3 & \cellcolor{drop!39}41.9 & \cellcolor{drop!55}39.5 & \cellcolor{drop!44}21.4 & \cellcolor{drop!10}63.9 & \cellcolor{drop!33}38.9 & \cellcolor{drop!19}63.0 & \cellcolor{drop!14}38.9 & \cellcolor{drop!31}53.3 & \cellcolor{drop!50}46.7 & 56.1 \\
DecompRCA & \cellcolor{drop!49}28.5 & \cellcolor{drop!46}34.4 & \cellcolor{drop!39}55.6 & \cellcolor{drop!42}23.8 & \cellcolor{drop!66}8.3 & \cellcolor{drop!42}30.6 & \cellcolor{drop!59}22.2 & \cellcolor{drop!53}0.0 & \cellcolor{drop!84}0.0 & \cellcolor{drop!83}13.3 & 36.9 \\
mABC$^{\S}$ & \cellcolor{drop!62}15.6 & \cellcolor{drop!65}16.1 & \cellcolor{drop!83}11.1 & \cellcolor{drop!58}7.1 & \cellcolor{drop!49}25.0 & \cellcolor{drop!35}37.5 & \cellcolor{drop!70}11.1 & \cellcolor{drop!44}8.3 & \cellcolor{drop!64}20.0 & \cellcolor{drop!97}0.0 & 20.1 \\
\specialrule{\lightrulewidth}{0pt}{0pt}
NSigma & \cellcolor{drop!38}38.9 & \cellcolor{drop!33}47.5 & \cellcolor{drop!35}59.3 & \cellcolor{drop!8}57.1 & \cellcolor{drop!49}25.0 & \cellcolor{drop!47}25.0 & \cellcolor{drop!70}11.1 & \cellcolor{drop!36}16.7 & \cellcolor{drop!84}0.0 & \cellcolor{drop!37}60.0 & 50.4 \\
TORAI$^{\ddagger}$ & \cellcolor{drop!44}33.3 & \cellcolor{drop!34}46.8 & \cellcolor{drop!31}63.0 & \cellcolor{drop!23}42.9 & \cellcolor{drop!49}25.0 & \cellcolor{drop!64}8.3 & \cellcolor{drop!70}11.1 & \cellcolor{drop!36}16.7 & \cellcolor{drop!84}0.0 & \cellcolor{drop!77}20.0 & 43.2 \\
SimpleRCA & \cellcolor{drop!44}33.3 & \cellcolor{drop!36}45.1 & \cellcolor{drop!39}55.6 & \cellcolor{drop!37}28.6 & \cellcolor{drop!57}16.7 & \cellcolor{drop!31}41.7 & \cellcolor{drop!37}44.4 & \cellcolor{drop!53}0.0 & \cellcolor{drop!84}0.0 & \cellcolor{drop!97}0.0 & 43.2 \\
BARO & \cellcolor{drop!55}22.2 & \cellcolor{drop!49}31.7 & \cellcolor{drop!50}44.4 & \cellcolor{drop!51}14.3 & \cellcolor{drop!66}8.3 & \cellcolor{drop!47}25.0 & \cellcolor{drop!59}22.2 & \cellcolor{drop!53}0.0 & \cellcolor{drop!84}0.0 & \cellcolor{drop!97}0.0 & 28.8 \\
CIRCA & \cellcolor{drop!59}17.8 & \cellcolor{drop!51}29.2 & \cellcolor{drop!65}29.6 & \cellcolor{drop!51}14.3 & \cellcolor{drop!74}0.0 & \cellcolor{drop!56}16.7 & \cellcolor{drop!48}33.3 & \cellcolor{drop!53}0.0 & \cellcolor{drop!64}20.0 & \cellcolor{drop!97}0.0 & 23.0 \\
FaultInsight$^{\ddagger}$ & \cellcolor{drop!58}18.9 & \cellcolor{drop!51}29.7 & \cellcolor{drop!66}28.4 & \cellcolor{drop!56}9.5 & \cellcolor{drop!55}19.4 & \cellcolor{drop!50}22.2 & \cellcolor{drop!59}22.2 & \cellcolor{drop!47}5.6 & \cellcolor{drop!78}6.7 & \cellcolor{drop!90}6.7 & 24.5 \\
RCD$^{\ddagger}$ & \cellcolor{drop!71}5.9 & \cellcolor{drop!70}10.5 & \cellcolor{drop!94}0.0 & \cellcolor{drop!63}2.4 & \cellcolor{drop!69}5.6 & \cellcolor{drop!64}8.3 & \cellcolor{drop!56}25.9 & \cellcolor{drop!53}0.0 & \cellcolor{drop!64}20.0 & \cellcolor{drop!97}0.0 & 7.7 \\
Random & \cellcolor{drop!72}5.5 & \cellcolor{drop!69}12.0 & \cellcolor{drop!87}7.1 & \cellcolor{drop!62}3.4 & \cellcolor{drop!66}8.2 & \cellcolor{drop!69}3.0 & \cellcolor{drop!73}8.3 & \cellcolor{drop!51}1.9 & \cellcolor{drop!79}5.8 & \cellcolor{drop!96}1.1 & 7.1 \\
\specialrule{\heavyrulewidth}{0pt}{\belowrulesep}\end{tabular}
\par\footnotesize $^{\dagger}$ Haiku not run. $^{\S}$ DeepSeek not run. $^{\ddagger}$ mean over three passes.
\end{table*}

\begin{table*}[t]
\caption{The five cost metrics of the three simple harnesses across models (setting 2, default effort; Astra at xhigh; first pass of each run): wall time per case, model calls per case, prompt and completion tokens per case, share of prompt tokens served from the provider's cache, and billed USD per case. Wall time excludes waits caused by the network (failed connections, read timeouts, reconnects). Calls = requests that reached the model (retries after connection failures during the 10-25 network outage are not counted). $^{*}$ = not billed per call (ChatGPT login); priced call by call at the OpenRouter list price, with the long-prompt tier (about twice the price) for calls above 272k input tokens, which is why the one-shot rows through Codex exceed the price of their mean token counts. MiMo: unfinished cases are not in the per-case means.}
\label{tab:costmetrics}\footnotesize\setlength{\tabcolsep}{5pt}
\begin{tabular}{llrrrrrr}
\toprule
Method & Model & \shortstack[r]{Time\\s / case} & \shortstack[r]{Calls\\/ case} & \shortstack[r]{Prompt\\tokens} & \shortstack[r]{Completion\\tokens} & \shortstack[r]{Cached\\share} & \shortstack[r]{Cost\\USD / case} \\
\midrule
One-shot & \raisebox{-0.25\height}{\includegraphics[height=1.1em]{logos/openai.pdf}}\,6-Luna & 46 & 1.0 & 235,644 & 3,949 & 0\% & 0.0522 \\
 & \raisebox{-0.25\height}{\includegraphics[height=1.1em]{logos/openai.pdf}}\,6.1-Sol & 64 & 1.0 & 241,783 & 1,392 & 21\% & 0.639$^{*}$ \\
 & \raisebox{-0.25\height}{\includegraphics[height=1.1em]{logos/anthropic.pdf}}\,5.5-Haiku & 55 & 1.0 & 269,460 & 10,731 & 0\% & 0.151 \\
 & \raisebox{-0.25\height}{\includegraphics[height=1.1em]{logos/deepseek.pdf}}\,4.1-Flash & 381 & 1.0 & 239,332 & 29,056 & 1\% & 0.0334 \\
 & \raisebox{-0.25\height}{\includegraphics[height=1.1em]{logos/xiaomi.pdf}}\,2.6-Pro & 1,037 & 1.0 & 293,483 & 57,216 & 14\% & 0.160 \\
 & \raisebox{-0.25\height}{\includegraphics[height=1.1em]{logos/openai.pdf}}\,6-Astra (xhigh) & 316 & 1.0 & 241,824 & 7,640 & 23\% & 3.56$^{*}$ \\
\addlinespace[2pt]
SQL agent & \raisebox{-0.25\height}{\includegraphics[height=1.1em]{logos/openai.pdf}}\,6-Luna & 134 & 19.7 & 381,235 & 6,268 & 90\% & 0.0112 \\
 & \raisebox{-0.25\height}{\includegraphics[height=1.1em]{logos/openai.pdf}}\,6.1-Sol & 45 & 8.9 & 90,160 & 2,260 & 76\% & 0.0827 \\
 & \raisebox{-0.25\height}{\includegraphics[height=1.1em]{logos/anthropic.pdf}}\,5.5-Haiku & 51 & 13.8 & 290,719 & 7,814 & 88\% & 0.0107 \\
 & \raisebox{-0.25\height}{\includegraphics[height=1.1em]{logos/deepseek.pdf}}\,4.1-Flash & 171 & 47.4 & 2,098,190 & 24,249 & 95\% & 0.0511 \\
 & \raisebox{-0.25\height}{\includegraphics[height=1.1em]{logos/xiaomi.pdf}}\,2.6-Pro & 997 & 46.3 & 2,628,405 & 38,658 & 93\% & 0.116 \\
\addlinespace[2pt]
Codex & \raisebox{-0.25\height}{\includegraphics[height=1.1em]{logos/openai.pdf}}\,6-Luna & 69 & 8.5 & 275,018 & 4,295 & 84\% & 0.00900$^{*}$ \\
 & \raisebox{-0.25\height}{\includegraphics[height=1.1em]{logos/openai.pdf}}\,6.1-Sol & 105 & 7.5 & 289,443 & 3,685 & 85\% & 0.149$^{*}$ \\
 & \raisebox{-0.25\height}{\includegraphics[height=1.1em]{logos/deepseek.pdf}}\,4.1-Flash & 574 & 34.0 & 2,552,729 & 80,595 & 90\% & 0.0549$^{*}$ \\
 & \raisebox{-0.25\height}{\includegraphics[height=1.1em]{logos/xiaomi.pdf}}\,2.6-Pro & 609 & 21.0 & 1,336,211 & 21,362 & 90\% & 0.0790$^{*}$ \\
 & \raisebox{-0.25\height}{\includegraphics[height=1.1em]{logos/openai.pdf}}\,6-Astra (xhigh) & 216 & 8.9 & 428,257 & 7,803 & 87\% & 1.32$^{*}$ \\
\addlinespace[2pt]
\bottomrule\end{tabular}\end{table*}

\section{Benchmark and method details}\label{app:details}
\paragraph{Benchmark.} Cases are converted to one format (entity and relation tables, metrics, traces, logs, events; UTC), with traces of the injection tooling removed. Of 542 screened cases, 348 pass the exclusion rules; the most frequent reasons are an unclean pre-fault period (69), no evidence (56), incomplete data (38) and a fault late in the window (37). LEMMA has no traces and hundreds of pods per scenario. RCAEval injects faults into only five services per system, so guessing among the injected services hits 20\%; against the full inventory our random-guess rate is far lower (Section~\ref{sec:benchmark}).

\subsection{Case construction}\label{app:cleaning}
Cases share entity and relation tables and UTC telemetry; labels and answer material are kept separate. Duplicate metric observations are collapsed, source attributes distinguish otherwise duplicate rows, and injection-tool entities, spans, logs and events are removed. Source fault windows are retained where available; RCA-100 adds 20 minutes before the alert, and sources giving only an injection time use an initial before-period and flag the remainder without exposing the onset.
Automatic checks cover label consistency, culprit coverage, fault timing, completeness, shortcuts and observable evidence. Claude Sonnet 5.5 reviews labels, check results, row counts and the twelve most-changed series; ITBench-SRE uses automatic checks only. Exclusions are contradictory labels (8), multiple faults outside the multi-fault scenario (20), a missing culprit or its data (14), no evidence (56), incomplete data (38), failure of the multi-fault pre-filter (5), an unclean before-period or nearby fault (69), and less than a third of the window remaining after the fault (37); rules overlap. Of 542 screened cases, 348 remain. RCAEval supplies three cases per sub-suite (27); other scenarios supply at least three where eligible. Remaining places diversify scenario, fault class, root-cause level, data kinds, volume and clue coverage, with fixed tie-breaking and seed, before method runs.

\subsection{One-shot evidence package}\label{app:package}
The prompt includes instructions, the question, recorded and flagged periods, all selected evidence and the scenario's answer template, with neutral item titles. S1 gives the incident record, inventory, topology, placement and coverage. S2 supplies moved metrics as raw samples: using the before-period median and robust spread (MAD $\times 1.4826$), a series needs a change score $\max(\text{median shift},0.6\times\text{largest departure})\geq4$ spreads, plus a median shift of at least 10\% or a departure of at least 20\% of the baseline level. Identical series are collapsed; unlabelled duplicate-row series are omitted. At most 120 series are selected in turns across moving entities and rendered on the source time grid, with missing samples marked.
S3 gives per-minute entry requests, errors, duration quantiles and caller--callee statistics, including missing downstream spans. S4 groups logs into masked templates: the 30 most-changed entities receive separate items, with up to eight templates and twelve raw lines per entity, prioritising errors, warnings and new templates. S5 groups Kubernetes events by object and reason (at most 60 groups). Constant series are counted only; series beyond the cap, full raw traces and logs, injection traces and answer material are omitted. The case study contains 136 items (about 75k tokens); no one-shot prompt was truncated.

\subsection{Method settings and hard groups}
\paragraph{Methods.} LLM methods run without generation parameters (except effort in Setting~3), one hour per case, first five answers scored. Interface fixes: shared raw-data tools for FoA, RCAgent and mABC; complete tool descriptions for FoA's code agent; UTC+8 times for RCA-Agent; modules missing from mABC's release, a loop limit, a cut at ``Observation'' and a format hint; resampling for TORAI; missing functions for FaultInsight. Setting differences: no temperatures; RCAgent's expert also reads metrics and traces (at most 30 steps); mABC's alert on the entry service's requests. The SQL agent differs from the reference agent of~\cite{fang2026openrca2} in temperature, an optional think tool, its answer format, and the topology files that every case folder carries.

\paragraph{Hardest groups.} On node-level faults the agents mention the true node in 85--90\% of runs but mostly blame a pod or service on it. Three AIOps~2022 cases fail for reasons in the data: no clear metric on the faulty node, a fault that leaves almost no trace, and an unlabelled burst on another node. No one-shot prompt was truncated; cases with all four data kinds have a median of 115k tokens.
\clearpage


\begin{thebibliography}{46}

\ifx \showCODEN    \undefined \def \showCODEN     #1{\unskip}     \fi
\ifx \showISBNx    \undefined \def \showISBNx     #1{\unskip}     \fi
\ifx \showISBNxiii \undefined \def \showISBNxiii  #1{\unskip}     \fi
\ifx \showISSN     \undefined \def \showISSN      #1{\unskip}     \fi
\ifx \showLCCN     \undefined \def \showLCCN      #1{\unskip}     \fi
\ifx \shownote     \undefined \def \shownote      #1{#1}          \fi
\ifx \showarticletitle \undefined \def \showarticletitle #1{#1}   \fi
\ifx \showURL      \undefined \def \showURL       {\relax}        \fi

\providecommand\bibfield[2]{#2}
\providecommand\bibinfo[2]{#2}
\providecommand\natexlab[1]{#1}
\providecommand\showeprint[2][]{arXiv:#2}
\makeatletter
\@ifundefined{NAT@parse@date}{}{\let\NAT@parse@date@orig\NAT@parse@date}
\@ifundefined{NAT@parse@date}{}{\def\NAT@parse@date#1#2#3#4#5#6@@{\NAT@parse@date@orig#1#2#3#4#5#6@@\def\NAT@tempyear{0000}\def\NAT@tempexlab{{?}}\ifx\NAT@year\NAT@tempyear\ifx\NAT@exlab\NAT@tempexlab\def\NAT@date{[n.\,d.]}\else\edef\NAT@date{[n.\,d.]\NAT@exlab}\fi\fi}}
\makeatother

\bibitem[{Anthropic}(2026)]
        {anthropic_claudecode}
\bibfield{author}{\bibinfo{person}{{Anthropic}}.}
  \bibinfo{year}{2026}\natexlab{}.
\newblock \bibinfo{title}{{Claude Code}}.
\newblock
\urldef\tempurl
\url{https://github.com/anthropics/claude-code}
\showURL{
\tempurl}
\newblock
\shownote{Software; accessed 2026-10-26}.

\bibitem[{Artificial Analysis}(2026)]
        {aa_index}
\bibfield{author}{\bibinfo{person}{{Artificial Analysis}}.}
  \bibinfo{year}{2026}\natexlab{}.
\newblock \bibinfo{title}{{Artificial Analysis Intelligence Index}}.
\newblock
\urldef\tempurl
\url{https://artificialanalysis.ai/methodology/intelligence-benchmarking}
\showURL{
\tempurl}
\newblock
\shownote{Version 4.3.2; accessed 2026-10-26}.

\bibitem[Beyer et~al\mbox{.}(2016)]
        {beyer2016site}
\bibfield{author}{\bibinfo{person}{Betsy Beyer}, \bibinfo{person}{Chris Jones},
  \bibinfo{person}{Jennifer Petoff}, {and} \bibinfo{person}{Niall~Richard
  Murphy}.} \bibinfo{year}{2016}\natexlab{}.
\newblock \bibinfo{booktitle}{\emph{Site reliability engineering: How Google
  runs production systems}}.
\newblock \bibinfo{publisher}{O'Reilly Media, Inc.}
\newblock

\bibitem[Bi et~al\mbox{.}(2024)]
        {bi2024faultinsight}
\bibfield{author}{\bibinfo{person}{Tingzhu Bi}, \bibinfo{person}{Zhang Yang},
  \bibinfo{person}{Yicheng Pan}, \bibinfo{person}{Yu Zhang},
  \bibinfo{person}{Meng Ma}, \bibinfo{person}{Xinrui Jiang},
  \bibinfo{person}{Linlin Han}, \bibinfo{person}{Feng Wang},
  \bibinfo{person}{Xian Liu}, {and} \bibinfo{person}{Ping Wang}.}
  \bibinfo{year}{2024}\natexlab{}.
\newblock \showarticletitle{{FaultInsight: Interpreting Hyperscale Data Center
  Host Faults}}. In \bibinfo{booktitle}{\emph{Proceedings of the 30th ACM
  SIGKDD Conference on Knowledge Discovery and Data Mining}}.
  \bibinfo{publisher}{Association for Computing Machinery},
  \bibinfo{pages}{141--152}.
\newblock
\href{https://doi.org/10.1145/3637528.3672051}{doi:\nolinkurl{10.1145/3637528.3672051}}

\bibitem[Bulji{\'c}(2026)]
        {buljic2026graph}
\bibfield{author}{\bibinfo{person}{Imad Bulji{\'c}}.}
  \bibinfo{year}{2026}\natexlab{}.
\newblock \bibinfo{title}{{Does Graph Structure Earn Its Place in Microservice
  Root-Cause Analysis? A Controlled Study on RCAEval, and What the Benchmark
  Was Really Measuring}}.
\newblock
\showeprint[arxiv]{2609.27069}
\urldef\tempurl
\url{https://arxiv.org/abs/2609.27069}
\showURL{
\tempurl}

\bibitem[Cai et~al\mbox{.}(2026)]
        {cai2026agenticopseval}
\bibfield{author}{\bibinfo{person}{Yuanhong Cai}, \bibinfo{person}{Xiaohui
  Nie}, \bibinfo{person}{Kanglin Yin}, \bibinfo{person}{Changhua Pei},
  \bibinfo{person}{Yongqian Sun}, \bibinfo{person}{Shenglin Zhang},
  \bibinfo{person}{Haibin Liu}, \bibinfo{person}{Guiyang Liu},
  \bibinfo{person}{Xidao Wen}, \bibinfo{person}{Fang Situ}, {and}
  \bibinfo{person}{Dan Pei}.} \bibinfo{year}{2026}\natexlab{}.
\newblock \bibinfo{title}{{A Multi-Dataset Benchmark for Evaluating LLM Agents
  in Microservice Failure Diagnosis}}.
\newblock
\showeprint[arxiv]{2606.29193}
\urldef\tempurl
\url{https://arxiv.org/abs/2606.29193}
\showURL{
\tempurl}

\bibitem[{China Computer Federation} et~al\mbox{.}(2022)]
        {aiops2022}
\bibfield{author}{\bibinfo{person}{{China Computer Federation}},
  \bibinfo{person}{{Tsinghua University}}, {and} \bibinfo{person}{{China
  Construction Bank}}.} \bibinfo{year}{2022}\natexlab{}.
\newblock \bibinfo{title}{{CCF International AIOps Challenge 2022}}.
\newblock
\urldef\tempurl
\url{https://www.aiops.cn/aiopschallenge/}
\showURL{
\tempurl}
\newblock
\shownote{Competition and dataset; accessed 2026-10-26}.

\bibitem[Dang et~al\mbox{.}(2019)]
        {dang2019aiops}
\bibfield{author}{\bibinfo{person}{Yingnong Dang}, \bibinfo{person}{Qingwei
  Lin}, {and} \bibinfo{person}{Peng Huang}.} \bibinfo{year}{2019}\natexlab{}.
\newblock \showarticletitle{AIOps: real-world challenges and research
  innovations}. In \bibinfo{booktitle}{\emph{2019 IEEE/ACM 41st International
  Conference on Software Engineering: Companion Proceedings (ICSE-Companion)}}.
  IEEE, \bibinfo{pages}{4--5}.
\newblock

\bibitem[Du et~al\mbox{.}(2025)]
        {du2025contextlength}
\bibfield{author}{\bibinfo{person}{Yufeng Du}, \bibinfo{person}{Minyang Tian},
  \bibinfo{person}{Srikanth Ronanki}, \bibinfo{person}{Subendhu Rongali},
  \bibinfo{person}{Sravan~Babu Bodapati}, \bibinfo{person}{Aram Galstyan},
  \bibinfo{person}{Azton Wells}, \bibinfo{person}{Roy Schwartz},
  \bibinfo{person}{Eliu~A Huerta}, {and} \bibinfo{person}{Hao Peng}.}
  \bibinfo{year}{2025}\natexlab{}.
\newblock \showarticletitle{{Context Length Alone Hurts LLM Performance Despite
  Perfect Retrieval}}. In \bibinfo{booktitle}{\emph{Findings of the Association
  for Computational Linguistics: EMNLP 2025}}. \bibinfo{publisher}{Association
  for Computational Linguistics}, \bibinfo{pages}{23281--23298}.
\newblock
\href{https://doi.org/10.18653/v1/2025.findings-emnlp.1264}{doi:\nolinkurl{10.18653/v1/2025.findings-emnlp.1264}}

\bibitem[Fang et~al\mbox{.}(2026a)]
        {fang2026openrca2}
\bibfield{author}{\bibinfo{person}{Aoyang Fang}, \bibinfo{person}{Yifan Yang},
  \bibinfo{person}{Jin'ao Shang}, \bibinfo{person}{Qisheng Lu},
  \bibinfo{person}{Junjielung Xu}, \bibinfo{person}{Rui Wang},
  \bibinfo{person}{Songhan Zhang}, \bibinfo{person}{Yuzhong Zhang},
  \bibinfo{person}{Boxi Yu}, {and} \bibinfo{person}{Pinjia He}.}
  \bibinfo{year}{2026}\natexlab{a}.
\newblock \bibinfo{title}{{OpenRCA 2.0: From Outcome Labels to Causal Process
  Supervision}}.
\newblock
\showeprint[arxiv]{2606.27154}
\urldef\tempurl
\url{https://arxiv.org/abs/2606.27154}
\showURL{
\tempurl}

\bibitem[Fang et~al\mbox{.}(2026b)]
        {fang2025rethinking}
\bibfield{author}{\bibinfo{person}{Aoyang Fang}, \bibinfo{person}{Songhan
  Zhang}, \bibinfo{person}{Yifan Yang}, \bibinfo{person}{Haotong Wu},
  \bibinfo{person}{Junjielong Xu}, \bibinfo{person}{Xuyang Wang},
  \bibinfo{person}{Rui Wang}, \bibinfo{person}{Manyi Wang},
  \bibinfo{person}{Qisheng Lu}, {and} \bibinfo{person}{Pinjia He}.}
  \bibinfo{year}{2026}\natexlab{b}.
\newblock \showarticletitle{{Rethinking the Evaluation of Microservice RCA with
  a Fault Propagation-Aware Benchmark}}.
\newblock \bibinfo{journal}{\emph{Proceedings of the ACM on Software
  Engineering}} \bibinfo{volume}{3}, \bibinfo{number}{FSE}
  (\bibinfo{year}{2026}), \bibinfo{pages}{1605--1627}.
\newblock
\href{https://doi.org/10.1145/3797100}{doi:\nolinkurl{10.1145/3797100}}

\bibitem[Huang et~al\mbox{.}(2026)]
        {huang2026opsharness}
\bibfield{author}{\bibinfo{person}{Haiyu Huang}, \bibinfo{person}{Jiewei Lyu},
  \bibinfo{person}{Zhihan Jiang}, \bibinfo{person}{Jinyang Liu},
  \bibinfo{person}{Xiao He}, \bibinfo{person}{Tieying Zhang},
  \bibinfo{person}{Wu Xiang}, {and} \bibinfo{person}{Michael~R. Lyu}.}
  \bibinfo{year}{2026}\natexlab{}.
\newblock \bibinfo{title}{{From General Agents to RCA Experts: A Self-Evolving
  Harness for Root Cause Analysis}}.
\newblock
\showeprint[arxiv]{2608.25661}
\urldef\tempurl
\url{https://arxiv.org/abs/2608.25661}
\showURL{
\tempurl}

\bibitem[Ikram et~al\mbox{.}(2022)]
        {ikram2022rcd}
\bibfield{author}{\bibinfo{person}{Azam Ikram}, \bibinfo{person}{Sarthak
  Chakraborty}, \bibinfo{person}{Subrata Mitra}, \bibinfo{person}{Shiv~Kumar
  Saini}, \bibinfo{person}{Saurabh Bagchi}, {and} \bibinfo{person}{Murat
  Kocaoglu}.} \bibinfo{year}{2022}\natexlab{}.
\newblock \showarticletitle{{Root Cause Analysis of Failures in Microservices
  through Causal Discovery}}. In \bibinfo{booktitle}{\emph{Advances in Neural
  Information Processing Systems}}, Vol.~\bibinfo{volume}{35}.
  \bibinfo{publisher}{Curran Associates, Inc.}, \bibinfo{pages}{31158--31170}.
\newblock
\href{https://doi.org/10.52202/068431-2259}{doi:\nolinkurl{10.52202/068431-2259}}

\bibitem[Jha et~al\mbox{.}(2025)]
        {jha2025itbench}
\bibfield{author}{\bibinfo{person}{Saurabh Jha}, \bibinfo{person}{Rohan~R.
  Arora}, \bibinfo{person}{Yuji Watanabe}, \bibinfo{person}{Takumi Yanagawa},
  \bibinfo{person}{Yinfang Chen}, \bibinfo{person}{Jackson Clark},
  \bibinfo{person}{Bhavya Bhavya}, \bibinfo{person}{Mudit Verma},
  \bibinfo{person}{Harshit Kumar}, \bibinfo{person}{Hirokuni Kitahara},
  \bibinfo{person}{Noah Zheutlin}, \bibinfo{person}{Saki Takano},
  \bibinfo{person}{Divya Pathak}, \bibinfo{person}{Felix George},
  \bibinfo{person}{Xinbo Wu}, \bibinfo{person}{Bekir~O Turkkan},
  \bibinfo{person}{Gerard Vanloo}, \bibinfo{person}{Michael Nidd},
  \bibinfo{person}{Ting Dai}, \bibinfo{person}{Oishik Chatterjee},
  \bibinfo{person}{Pranjal Gupta}, \bibinfo{person}{Suranjana Samanta},
  \bibinfo{person}{Pooja Aggarwal}, \bibinfo{person}{Rong Lee},
  \bibinfo{person}{Jae-Wook Ahn}, \bibinfo{person}{Debanjana Kar},
  \bibinfo{person}{Amit Paradkar}, \bibinfo{person}{Yu Deng},
  \bibinfo{person}{Pratibha Moogi}, \bibinfo{person}{Prateeti Mohapatra},
  \bibinfo{person}{Naoki Abe}, \bibinfo{person}{Chandrasekhar Narayanaswami},
  \bibinfo{person}{Tianyin Xu}, \bibinfo{person}{Lav~R. Varshney},
  \bibinfo{person}{Ruchi Mahindru}, \bibinfo{person}{Anca Sailer},
  \bibinfo{person}{Laura Shwartz}, \bibinfo{person}{Daby Sow},
  \bibinfo{person}{Nicholas C.~M. Fuller}, {and} \bibinfo{person}{Ruchir
  Puri}.} \bibinfo{year}{2025}\natexlab{}.
\newblock \showarticletitle{{ITBench: Evaluating AI Agents across Diverse
  Real-World IT Automation Tasks}}. In \bibinfo{booktitle}{\emph{Proceedings of
  the 42nd International Conference on Machine Learning}}
  \emph{(\bibinfo{series}{Proceedings of Machine Learning Research},
  Vol.~\bibinfo{volume}{267})}. \bibinfo{publisher}{PMLR},
  \bibinfo{pages}{27134--27197}.
\newblock
\urldef\tempurl
\url{https://proceedings.mlr.press/v267/jha25a.html}
\showURL{
\tempurl}

\bibitem[Kim et~al\mbox{.}(2026)]
        {kim2026whyfail}
\bibfield{author}{\bibinfo{person}{Taeyoon Kim}, \bibinfo{person}{Woohyeok
  Park}, \bibinfo{person}{Hoyeong Yun}, {and} \bibinfo{person}{Kyungyong Lee}.}
  \bibinfo{year}{2026}\natexlab{}.
\newblock \showarticletitle{{Why Do AI Agents Systematically Fail at Cloud Root
  Cause Analysis?}}. In \bibinfo{booktitle}{\emph{Cloud Intelligence/AIOps
  Workshop}}.
\newblock
\urldef\tempurl
\url{https://cloudintelligenceworkshop.org/papers/aiops26-Kim.pdf}
\showURL{
\tempurl}

\bibitem[Laban et~al\mbox{.}(2026)]
        {laban2025lost}
\bibfield{author}{\bibinfo{person}{Philippe Laban}, \bibinfo{person}{Hiroaki
  Hayashi}, \bibinfo{person}{Yingbo Zhou}, {and} \bibinfo{person}{Jennifer
  Neville}.} \bibinfo{year}{2026}\natexlab{}.
\newblock \showarticletitle{{LLMs Get Lost In Multi-Turn Conversation}}. In
  \bibinfo{booktitle}{\emph{The Fourteenth International Conference on Learning
  Representations}}.
\newblock
\urldef\tempurl
\url{https://openreview.net/forum?id=VKGTGGcwl6}
\showURL{
\tempurl}

\bibitem[Li et~al\mbox{.}(2022)]
        {li2022circa}
\bibfield{author}{\bibinfo{person}{Mingjie Li}, \bibinfo{person}{Zeyan Li},
  \bibinfo{person}{Kanglin Yin}, \bibinfo{person}{Xiaohui Nie},
  \bibinfo{person}{Wenchi Zhang}, \bibinfo{person}{Kaixin Sui}, {and}
  \bibinfo{person}{Dan Pei}.} \bibinfo{year}{2022}\natexlab{}.
\newblock \showarticletitle{{Causal Inference-Based Root Cause Analysis for
  Online Service Systems with Intervention Recognition}}. In
  \bibinfo{booktitle}{\emph{Proceedings of the 28th ACM SIGKDD Conference on
  Knowledge Discovery and Data Mining}}. \bibinfo{publisher}{Association for
  Computing Machinery}, \bibinfo{pages}{3230--3240}.
\newblock
\href{https://doi.org/10.1145/3534678.3539041}{doi:\nolinkurl{10.1145/3534678.3539041}}

\bibitem[Lin et~al\mbox{.}(2018)]
        {lin2018microscope}
\bibfield{author}{\bibinfo{person}{JinJin Lin}, \bibinfo{person}{Pengfei Chen},
  {and} \bibinfo{person}{Zibin Zheng}.} \bibinfo{year}{2018}\natexlab{}.
\newblock \showarticletitle{Microscope: Pinpoint performance issues with causal
  graphs in micro-service environments}. In
  \bibinfo{booktitle}{\emph{International Conference on Service-Oriented
  Computing}}. Springer, \bibinfo{pages}{3--20}.
\newblock

\bibitem[Lu et~al\mbox{.}(2026)]
        {lu2026beyond}
\bibfield{author}{\bibinfo{person}{Qisheng Lu}, \bibinfo{person}{Aoyang Fang},
  \bibinfo{person}{Junjielong Xu}, \bibinfo{person}{Jin'ao Shang},
  \bibinfo{person}{Songhan Zhang}, \bibinfo{person}{Yifan Yang},
  \bibinfo{person}{Xiaochuan Yan}, {and} \bibinfo{person}{Pinjia He}.}
  \bibinfo{year}{2026}\natexlab{}.
\newblock \bibinfo{title}{{Beyond Fault Localization: A Trajectory-Level Study
  of LLM Agents for Microservice Root Cause Analysis}}.
\newblock
\showeprint[arxiv]{2608.21310}
\urldef\tempurl
\url{https://arxiv.org/abs/2608.21310}
\showURL{
\tempurl}

\bibitem[Ma et~al\mbox{.}(2020)]
        {ma2020automap}
\bibfield{author}{\bibinfo{person}{Meng Ma}, \bibinfo{person}{Jingmin Xu},
  \bibinfo{person}{Yuan Wang}, \bibinfo{person}{Pengfei Chen},
  \bibinfo{person}{Zonghua Zhang}, {and} \bibinfo{person}{Ping Wang}.}
  \bibinfo{year}{2020}\natexlab{}.
\newblock \showarticletitle{AutoMAP: Diagnose Your Microservice-based Web
  Applications Automatically}. In \bibinfo{booktitle}{\emph{{WWW} '20: The Web
  Conference 2020, Taipei, Taiwan, April 20-24, 2020}}.
  \bibinfo{publisher}{{ACM} / {IW3C2}}, \bibinfo{pages}{246--258}.
\newblock

\bibitem[Muhammad et~al\mbox{.}(2026)]
        {muhammad2026decomprca}
\bibfield{author}{\bibinfo{person}{Hada~Melino Muhammad}, \bibinfo{person}{Luan
  Pham}, \bibinfo{person}{Laure Barri{\`e}re}, \bibinfo{person}{Sachin Shetty},
  \bibinfo{person}{Leonardo Pulga}, {and} \bibinfo{person}{Flora~D. Salim}.}
  \bibinfo{year}{2026}\natexlab{}.
\newblock \bibinfo{title}{{Where Root Cause Analysis Fails: A
  Retrieval-Reranking Decomposition}}.
\newblock
\showeprint[arxiv]{2609.36686}
\urldef\tempurl
\url{https://arxiv.org/abs/2609.36686}
\showURL{
\tempurl}
\newblock
\shownote{Accepted at NeurIPS 2026, Evaluations and Datasets Track; to appear}.

\bibitem[Naakka et~al\mbox{.}(2027)]
        {naakka2026latsrca}
\bibfield{author}{\bibinfo{person}{Alexander Naakka}, \bibinfo{person}{Yuqing
  Wang}, {and} \bibinfo{person}{Mika~V. M{\"a}ntyl{\"a}}.}
  \bibinfo{year}{2027}\natexlab{}.
\newblock \showarticletitle{{LATS-RCA: Language Agent Tree Search for Root
  Cause Analysis in Microservices}}. In \bibinfo{booktitle}{\emph{Software
  Engineering and Advanced Applications}} \emph{(\bibinfo{series}{Lecture Notes
  in Computer Science}, Vol.~\bibinfo{volume}{16863})}.
  \bibinfo{publisher}{Springer Nature Switzerland}, \bibinfo{pages}{196--212}.
\newblock
\href{https://doi.org/10.1007/978-3-032-36590-3_14}{doi:\nolinkurl{10.1007/978-3-032-36590-3_14}}
\newblock
\shownote{SEAA 2026; first published online 27 August 2026}.

\bibitem[{OpenAI}(2026)]
        {openai_codex}
\bibfield{author}{\bibinfo{person}{{OpenAI}}.} \bibinfo{year}{2026}\natexlab{}.
\newblock \bibinfo{title}{{Codex CLI}}.
\newblock
\urldef\tempurl
\url{https://github.com/openai/codex}
\showURL{
\tempurl}
\newblock
\shownote{Software; accessed 2026-10-26}.

\bibitem[Pan et~al\mbox{.}(2021)]
        {pan2021dycause}
\bibfield{author}{\bibinfo{person}{Yicheng Pan}, \bibinfo{person}{Meng Ma},
  \bibinfo{person}{Xinrui Jiang}, {and} \bibinfo{person}{Ping Wang}.}
  \bibinfo{year}{2021}\natexlab{}.
\newblock \showarticletitle{{Faster, deeper, easier: crowdsourcing diagnosis of
  microservice kernel failure from user space}}. In
  \bibinfo{booktitle}{\emph{Proceedings of the 30th ACM SIGSOFT International
  Symposium on Software Testing and Analysis}}. \bibinfo{publisher}{Association
  for Computing Machinery}, \bibinfo{pages}{646--657}.
\newblock
\href{https://doi.org/10.1145/3460319.3464805}{doi:\nolinkurl{10.1145/3460319.3464805}}

\bibitem[Pei et~al\mbox{.}(2025)]
        {pei2025foa}
\bibfield{author}{\bibinfo{person}{Changhua Pei}, \bibinfo{person}{Zexin Wang},
  \bibinfo{person}{Fengrui Liu}, \bibinfo{person}{Zeyan Li},
  \bibinfo{person}{Yang Liu}, \bibinfo{person}{Xiao He}, \bibinfo{person}{Rong
  Kang}, \bibinfo{person}{Tieying Zhang}, \bibinfo{person}{Jianjun Chen},
  \bibinfo{person}{Jianhui Li}, \bibinfo{person}{Gaogang Xie}, {and}
  \bibinfo{person}{Dan Pei}.} \bibinfo{year}{2025}\natexlab{}.
\newblock \showarticletitle{{Flow-of-Action: SOP Enhanced LLM-Based Multi-Agent
  System for Root Cause Analysis}}. In \bibinfo{booktitle}{\emph{Companion
  Proceedings of the ACM on Web Conference 2025}}.
  \bibinfo{publisher}{Association for Computing Machinery},
  \bibinfo{pages}{422--431}.
\newblock
\href{https://doi.org/10.1145/3701716.3715225}{doi:\nolinkurl{10.1145/3701716.3715225}}

\bibitem[Pham et~al\mbox{.}(2024a)]
        {pham2024baro}
\bibfield{author}{\bibinfo{person}{Luan Pham}, \bibinfo{person}{Huong Ha},
  {and} \bibinfo{person}{Hongyu Zhang}.} \bibinfo{year}{2024}\natexlab{a}.
\newblock \showarticletitle{{BARO: Robust Root Cause Analysis for Microservices
  via Multivariate Bayesian Online Change Point Detection}}.
\newblock \bibinfo{journal}{\emph{Proceedings of the ACM on Software
  Engineering}} \bibinfo{volume}{1}, \bibinfo{number}{FSE}
  (\bibinfo{year}{2024}), \bibinfo{pages}{2214--2237}.
\newblock
\href{https://doi.org/10.1145/3660805}{doi:\nolinkurl{10.1145/3660805}}

\bibitem[Pham et~al\mbox{.}(2024b)]
        {pham2024howfar}
\bibfield{author}{\bibinfo{person}{Luan Pham}, \bibinfo{person}{Huong Ha},
  {and} \bibinfo{person}{Hongyu Zhang}.} \bibinfo{year}{2024}\natexlab{b}.
\newblock \showarticletitle{{Root Cause Analysis for Microservice System based
  on Causal Inference: How Far Are We?}}. In
  \bibinfo{booktitle}{\emph{Proceedings of the 39th IEEE/ACM International
  Conference on Automated Software Engineering}}.
  \bibinfo{publisher}{Association for Computing Machinery},
  \bibinfo{pages}{706--715}.
\newblock
\href{https://doi.org/10.1145/3691620.3695065}{doi:\nolinkurl{10.1145/3691620.3695065}}

\bibitem[Pham et~al\mbox{.}(2026)]
        {pham2026torai}
\bibfield{author}{\bibinfo{person}{Luan Pham}, \bibinfo{person}{Huong Ha},
  \bibinfo{person}{Xiuzhen Zhang}, {and} \bibinfo{person}{Hongyu Zhang}.}
  \bibinfo{year}{2026}\natexlab{}.
\newblock \showarticletitle{{TORAI: Multi-source Root Cause Analysis for Blind
  Spots in Microservice Service Call Graph}}.
\newblock \bibinfo{journal}{\emph{Proceedings of the ACM on Software
  Engineering}} \bibinfo{volume}{3}, \bibinfo{number}{FSE}
  (\bibinfo{year}{2026}), \bibinfo{pages}{2928--2951}.
\newblock
\href{https://doi.org/10.1145/3808137}{doi:\nolinkurl{10.1145/3808137}}

\bibitem[Pham et~al\mbox{.}(2025)]
        {pham2025rcaeval}
\bibfield{author}{\bibinfo{person}{Luan Pham}, \bibinfo{person}{Hongyu Zhang},
  \bibinfo{person}{Huong Ha}, \bibinfo{person}{Flora Salim}, {and}
  \bibinfo{person}{Xiuzhen Zhang}.} \bibinfo{year}{2025}\natexlab{}.
\newblock \showarticletitle{{RCAEval: A Benchmark for Root Cause Analysis of
  Microservice Systems with Telemetry Data}}. In
  \bibinfo{booktitle}{\emph{Companion Proceedings of the ACM on Web Conference
  2025}}. \bibinfo{publisher}{Association for Computing Machinery},
  \bibinfo{pages}{777--780}.
\newblock
\href{https://doi.org/10.1145/3701716.3715290}{doi:\nolinkurl{10.1145/3701716.3715290}}

\bibitem[Riddell et~al\mbox{.}(2026)]
        {riddell2026stalled}
\bibfield{author}{\bibinfo{person}{Evelien Riddell}, \bibinfo{person}{James
  Riddell}, \bibinfo{person}{Gengyi Sun}, \bibinfo{person}{Micha{\l}
  Antkiewicz}, {and} \bibinfo{person}{Krzysztof Czarnecki}.}
  \bibinfo{year}{2026}\natexlab{}.
\newblock \showarticletitle{{Stalled, Biased, and Confused: Uncovering
  Reasoning Failures in LLMs for Cloud-Based Root Cause Analysis}}. In
  \bibinfo{booktitle}{\emph{Proceedings of the 2026 IEEE/ACM Third
  International Conference on AI Foundation Models and Software Engineering}}.
  \bibinfo{publisher}{Association for Computing Machinery},
  \bibinfo{pages}{172--183}.
\newblock
\href{https://doi.org/10.1145/3793655.3793732}{doi:\nolinkurl{10.1145/3793655.3793732}}

\bibitem[Sebin et~al\mbox{.}(2026)]
        {itbench_aa}
\bibfield{author}{\bibinfo{person}{Ayhan Sebin}, \bibinfo{person}{Saurabh Jha},
  {and} \bibinfo{person}{Rohan Arora}.} \bibinfo{year}{2026}\natexlab{}.
\newblock \bibinfo{title}{{ITBench-AA: Frontier Models Score Below 50\% on the
  First Benchmark for Agentic Enterprise IT Tasks --- by Artificial Analysis
  and IBM}}.
\newblock
\urldef\tempurl
\url{https://huggingface.co/blog/ibm-research/itbench-aa}
\showURL{
\tempurl}
\newblock
\shownote{Hugging Face community article; accessed 2026-10-26}.

\bibitem[Shan et~al\mbox{.}(2019)]
        {shan2019ediagnosis}
\bibfield{author}{\bibinfo{person}{Huasong Shan}, \bibinfo{person}{Yuan Chen},
  \bibinfo{person}{Haifeng Liu}, \bibinfo{person}{Yunpeng Zhang},
  \bibinfo{person}{Xiao Xiao}, \bibinfo{person}{Xiaofeng He},
  \bibinfo{person}{Min Li}, {and} \bibinfo{person}{Wei Ding}.}
  \bibinfo{year}{2019}\natexlab{}.
\newblock \showarticletitle{{$\epsilon$-Diagnosis: Unsupervised and Real-time
  Diagnosis of Small-window Long-tail Latency in Large-scale Microservice
  Platforms}}. In \bibinfo{booktitle}{\emph{The World Wide Web Conference}}.
  \bibinfo{publisher}{Association for Computing Machinery},
  \bibinfo{pages}{3215--3222}.
\newblock
\href{https://doi.org/10.1145/3308558.3313653}{doi:\nolinkurl{10.1145/3308558.3313653}}

\bibitem[Sutton(2019)]
        {sutton2019bitter}
\bibfield{author}{\bibinfo{person}{Richard~S. Sutton}.}
  \bibinfo{year}{2019}\natexlab{}.
\newblock \bibinfo{title}{{The Bitter Lesson}}.
\newblock
\urldef\tempurl
\url{http://www.incompleteideas.net/IncIdeas/BitterLesson.html}
\showURL{
\tempurl}
\newblock
\shownote{Online essay; 13 March 2019; accessed 2026-10-26}.

\bibitem[Tang et~al\mbox{.}(2025)]
        {tang2025microrcaagent}
\bibfield{author}{\bibinfo{person}{Pan Tang}, \bibinfo{person}{Shixiang Tang},
  \bibinfo{person}{Huanqi Pu}, \bibinfo{person}{Zhiqing Miao}, {and}
  \bibinfo{person}{Zhixing Wang}.} \bibinfo{year}{2025}\natexlab{}.
\newblock \bibinfo{title}{{MicroRCA-Agent: Microservice Root Cause Analysis
  Method Based on Large Language Model Agents}}.
\newblock
\showeprint[arxiv]{2509.15635}
\urldef\tempurl
\url{https://arxiv.org/abs/2509.15635}
\showURL{
\tempurl}

\bibitem[Wang et~al\mbox{.}(2018)]
        {wang2018cloudranger}
\bibfield{author}{\bibinfo{person}{Ping Wang}, \bibinfo{person}{Jingmin Xu},
  \bibinfo{person}{Meng Ma}, \bibinfo{person}{Weilan Lin},
  \bibinfo{person}{Disheng Pan}, \bibinfo{person}{Yuan Wang}, {and}
  \bibinfo{person}{Pengfei Chen}.} \bibinfo{year}{2018}\natexlab{}.
\newblock \showarticletitle{CloudRanger: Root Cause Identification for Cloud
  Native Systems}. In \bibinfo{booktitle}{\emph{18th {IEEE/ACM} International
  Symposium on Cluster, Cloud and Grid Computing, {CCGRID} 2018, Washington,
  DC, USA, May 1-4, 2018}}. \bibinfo{publisher}{{IEEE} Computer Society},
  \bibinfo{pages}{492--502}.
\newblock

\bibitem[Wang et~al\mbox{.}(2026)]
        {wang2026evirca}
\bibfield{author}{\bibinfo{person}{Yuhao Wang}, \bibinfo{person}{Zhen Qin},
  \bibinfo{person}{Xingliang Wang}, \bibinfo{person}{Guochang Li},
  \bibinfo{person}{Weize Li}, {and} \bibinfo{person}{Shuiguang Deng}.}
  \bibinfo{year}{2026}\natexlab{}.
\newblock \bibinfo{title}{{EviRCA: Decoupling Evidence Extraction from
  Reasoning for Microservice Root-Cause Analysis}}.
\newblock
\showeprint[arxiv]{2609.19825}
\urldef\tempurl
\url{https://arxiv.org/abs/2609.19825}
\showURL{
\tempurl}

\bibitem[Wang et~al\mbox{.}(2024)]
        {wang2024rcagent}
\bibfield{author}{\bibinfo{person}{Zefan Wang}, \bibinfo{person}{Zichuan Liu},
  \bibinfo{person}{Yingying Zhang}, \bibinfo{person}{Aoxiao Zhong},
  \bibinfo{person}{Jihong Wang}, \bibinfo{person}{Fengbin Yin},
  \bibinfo{person}{Lunting Fan}, \bibinfo{person}{Lingfei Wu}, {and}
  \bibinfo{person}{Qingsong Wen}.} \bibinfo{year}{2024}\natexlab{}.
\newblock \showarticletitle{{RCAgent: Cloud Root Cause Analysis by Autonomous
  Agents with Tool-Augmented Large Language Models}}. In
  \bibinfo{booktitle}{\emph{Proceedings of the 33rd ACM International
  Conference on Information and Knowledge Management}}.
  \bibinfo{publisher}{Association for Computing Machinery},
  \bibinfo{pages}{4966--4974}.
\newblock
\href{https://doi.org/10.1145/3627673.3680016}{doi:\nolinkurl{10.1145/3627673.3680016}}

\bibitem[Wen et~al\mbox{.}(2026)]
        {rca100_2026}
\bibfield{author}{\bibinfo{person}{Xidao Wen}, \bibinfo{person}{Haibin Liu},
  \bibinfo{person}{Guiyang Liu}, \bibinfo{person}{Cheng Zhang},
  \bibinfo{person}{Fang Situ}, {and} \bibinfo{person}{Qi Zhou}.}
  \bibinfo{year}{2026}\natexlab{}.
\newblock \bibinfo{title}{{RCA-100: A Chain-Reasoning Benchmark for Root Cause
  Analysis on Cloud-Native Microservices}}.
\newblock
\urldef\tempurl
\url{https://www.aiops.cn/gitlab/aiops-live-benchmark/agenticopseval/-/tree/main/RCA100}
\showURL{
\tempurl}
\newblock
\shownote{Dataset; 103 chaos-drill incidents; CC BY-NC-SA 4.0; accessed
  2026-10-26}.

\bibitem[Wu et~al\mbox{.}(2020)]
        {wu2020microrca}
\bibfield{author}{\bibinfo{person}{Li Wu}, \bibinfo{person}{Johan Tordsson},
  \bibinfo{person}{Erik Elmroth}, {and} \bibinfo{person}{Odej Kao}.}
  \bibinfo{year}{2020}\natexlab{}.
\newblock \showarticletitle{{MicroRCA: Root Cause Localization of Performance
  Issues in Microservices}}. In \bibinfo{booktitle}{\emph{NOMS 2020 - 2020
  IEEE/IFIP Network Operations and Management Symposium}}.
  \bibinfo{publisher}{IEEE}, \bibinfo{pages}{1--9}.
\newblock
\href{https://doi.org/10.1109/noms47738.2020.9110353}{doi:\nolinkurl{10.1109/noms47738.2020.9110353}}

\bibitem[Xia et~al\mbox{.}(2025)]
        {xia2024agentless}
\bibfield{author}{\bibinfo{person}{Chunqiu~Steven Xia}, \bibinfo{person}{Yinlin
  Deng}, \bibinfo{person}{Soren Dunn}, {and} \bibinfo{person}{Lingming Zhang}.}
  \bibinfo{year}{2025}\natexlab{}.
\newblock \showarticletitle{{Demystifying LLM-Based Software Engineering
  Agents}}.
\newblock \bibinfo{journal}{\emph{Proceedings of the ACM on Software
  Engineering}} \bibinfo{volume}{2}, \bibinfo{number}{FSE}
  (\bibinfo{year}{2025}), \bibinfo{pages}{801--824}.
\newblock
\href{https://doi.org/10.1145/3715754}{doi:\nolinkurl{10.1145/3715754}}

\bibitem[Xu et~al\mbox{.}(2025)]
        {xu2025openrca}
\bibfield{author}{\bibinfo{person}{Junjielong Xu}, \bibinfo{person}{Qinan
  Zhang}, \bibinfo{person}{Zhiqing Zhong}, \bibinfo{person}{Shilin He},
  \bibinfo{person}{Chaoyun Zhang}, \bibinfo{person}{Qingwei Lin},
  \bibinfo{person}{Dan Pei}, \bibinfo{person}{Pinjia He},
  \bibinfo{person}{Dongmei Zhang}, {and} \bibinfo{person}{Qi Zhang}.}
  \bibinfo{year}{2025}\natexlab{}.
\newblock \showarticletitle{{OpenRCA: Can Large Language Models Locate the Root
  Cause of Software Failures?}}. In \bibinfo{booktitle}{\emph{International
  Conference on Learning Representations}}. \bibinfo{pages}{84562--84590}.
\newblock
\urldef\tempurl
\url{https://proceedings.iclr.cc/paper_files/paper/2025/hash/d29b8d53678015079e1d245c023e49d2-Abstract-Conference.html}
\showURL{
\tempurl}

\bibitem[Yu et~al\mbox{.}(2021)]
        {yu2021microrank}
\bibfield{author}{\bibinfo{person}{Guangba Yu}, \bibinfo{person}{Pengfei Chen},
  \bibinfo{person}{Hongyang Chen}, \bibinfo{person}{Zijie Guan},
  \bibinfo{person}{Zicheng Huang}, \bibinfo{person}{Linxiao Jing},
  \bibinfo{person}{Tianjun Weng}, \bibinfo{person}{Xinmeng Sun}, {and}
  \bibinfo{person}{Xiaoyun Li}.} \bibinfo{year}{2021}\natexlab{}.
\newblock \showarticletitle{{MicroRank: End-to-End Latency Issue Localization
  with Extended Spectrum Analysis in Microservice Environments}}. In
  \bibinfo{booktitle}{\emph{Proceedings of the Web Conference 2021}}.
  \bibinfo{publisher}{Association for Computing Machinery},
  \bibinfo{pages}{3087--3098}.
\newblock
\href{https://doi.org/10.1145/3442381.3449905}{doi:\nolinkurl{10.1145/3442381.3449905}}

\bibitem[Zhang et~al\mbox{.}(2026)]
        {rclagentv2}
\bibfield{author}{\bibinfo{person}{Lingzhe Zhang}, \bibinfo{person}{Tong Jia},
  \bibinfo{person}{Kangjin Wang}, \bibinfo{person}{Chiming Duan},
  \bibinfo{person}{Minghua He}, \bibinfo{person}{Rongqian Wang},
  \bibinfo{person}{Xi Peng}, \bibinfo{person}{Meiling Wang},
  \bibinfo{person}{Gong Zhang}, \bibinfo{person}{Renhai Chen}, {and}
  \bibinfo{person}{Ying Li}.} \bibinfo{year}{2026}\natexlab{}.
\newblock \showarticletitle{{Toward In-Depth Root Cause Localization for
  Microservices With Multi-Agent Recursion-of-Thought}}.
\newblock \bibinfo{journal}{\emph{IEEE Transactions on Dependable and Secure
  Computing}} \bibinfo{volume}{23}, \bibinfo{number}{5} (\bibinfo{year}{2026}),
  \bibinfo{pages}{10702--10719}.
\newblock
\href{https://doi.org/10.1109/tdsc.2026.3706572}{doi:\nolinkurl{10.1109/tdsc.2026.3706572}}

\bibitem[Zhang et~al\mbox{.}(2024)]
        {zhang2024mabc}
\bibfield{author}{\bibinfo{person}{Wei Zhang}, \bibinfo{person}{Hongcheng Guo},
  \bibinfo{person}{Jian Yang}, \bibinfo{person}{Zhoujin Tian},
  \bibinfo{person}{Yi Zhang}, \bibinfo{person}{Chaoran Yan},
  \bibinfo{person}{Zhoujun Li}, \bibinfo{person}{Tongliang Li},
  \bibinfo{person}{Xu Shi}, \bibinfo{person}{Liangfan Zheng}, {and}
  \bibinfo{person}{Bo Zhang}.} \bibinfo{year}{2024}\natexlab{}.
\newblock \showarticletitle{{mABC: Multi-Agent Blockchain-inspired
  Collaboration for Root Cause Analysis in Micro-Services Architecture}}. In
  \bibinfo{booktitle}{\emph{Findings of the Association for Computational
  Linguistics: EMNLP 2024}}. \bibinfo{publisher}{Association for Computational
  Linguistics}, \bibinfo{pages}{4017--4033}.
\newblock
\href{https://doi.org/10.18653/v1/2024.findings-emnlp.232}{doi:\nolinkurl{10.18653/v1/2024.findings-emnlp.232}}

\bibitem[Zheng et~al\mbox{.}(2026)]
        {zheng2024lemmarca}
\bibfield{author}{\bibinfo{person}{Lecheng Zheng}, \bibinfo{person}{Zhengzhang
  Chen}, \bibinfo{person}{Dongjie Wang}, \bibinfo{person}{Chengyuan Deng},
  \bibinfo{person}{Reon Matsuoka}, {and} \bibinfo{person}{Haifeng Chen}.}
  \bibinfo{year}{2026}\natexlab{}.
\newblock \showarticletitle{{LEMMA-RCA: A Large Multi-modal Multi-domain
  Dataset for Root Cause Analysis}}. In \bibinfo{booktitle}{\emph{Proceedings
  of the 35th ACM International Conference on Information and Knowledge
  Management}}.
\newblock
\href{https://doi.org/10.1145/3799682.3840176}{doi:\nolinkurl{10.1145/3799682.3840176}}
\newblock
\shownote{To appear}.

\bibitem[Zhou et~al\mbox{.}(2021)]
        {zhou2018fault}
\bibfield{author}{\bibinfo{person}{Xiang Zhou}, \bibinfo{person}{Xin Peng},
  \bibinfo{person}{Tao Xie}, \bibinfo{person}{Jun Sun}, \bibinfo{person}{Chao
  Ji}, \bibinfo{person}{Wenhai Li}, {and} \bibinfo{person}{Dan Ding}.}
  \bibinfo{year}{2021}\natexlab{}.
\newblock \showarticletitle{Fault Analysis and Debugging of Microservice
  Systems: Industrial Survey, Benchmark System, and Empirical Study}.
\newblock \bibinfo{journal}{\emph{{IEEE} Trans. Software Eng.}}
  \bibinfo{volume}{47}, \bibinfo{number}{2} (\bibinfo{year}{2021}),
  \bibinfo{pages}{243--260}.
\newblock

\end{thebibliography}
\end{document}